\let\origvec\vec
\documentclass[twocolumn]{svjour3}

\let\vec\origvec
\smartqed  

\usepackage{eqparbox}
\usepackage[left=1.8cm,right=1.8cm,top=2.0cm,bottom=2.7cm]{geometry}

\usepackage[numbers,sort]{natbib}
\bibpunct[,]{(}{)}{;}{a}{}{,}
\renewcommand{\cite}[1]{\textcolor{blue}{\citep{#1}}}

\usepackage{graphics} 
\usepackage{graphicx}
\usepackage{amsmath} 
\usepackage{amssymb}  
\usepackage{color}
\usepackage{hyperref}

\usepackage{multirow}
\usepackage{algorithm,algcompatible}
\usepackage{bm}
\usepackage[tight,footnotesize]{subfigure}
\usepackage{multicol}

\algnewcommand\INPUT{\item[\textbf{Input:}]}%
\algnewcommand\OUTPUT{\item[\textbf{Output:}]}%

\newcommand{\etal}{~\textit{et al}.}
\newcommand{\uvec}[1]{\boldsymbol{\hat{\textbf{#1}}}}

\newcommand{\figref}[1]{Fig.~\ref{#1}}
\newcommand{\tabref}[1]{Tab.~\ref{#1}}
\newcommand{\eqnref}[1]{Eqn.~\ref{#1}}
\newcommand{\secref}[1]{Sec.~\ref{#1}}
\newcommand{\algmref}[1]{Alg.~\ref{#1}}

\journalname{UNDER REVIEW}

\begin{document}

\title{DF-VO: What Should Be Learnt for Visual Odometry?}
\author{Huangying Zhan, Chamara Saroj Weerasekera, Jia-Wang Bian, Ravi Garg, Ian Reid
\institute{
All authors are with the University of Adelaide, and Australian Centre for Robotic Vision}
}

\authorrunning{Zhan et al.} 


\maketitle
\begin{abstract}
Multi-view geometry-based methods dominate the last few decades in monocular Visual Odometry for their superior performance,
while they have been vulnerable to dynamic and low-texture scenes. 
More importantly, monocular methods suffer from \textit{scale-drift} issue, i.e., errors accumulate over time.
Recent studies show that deep neural networks can learn scene depths and relative camera in a self-supervised manner without acquiring ground truth labels.
More surprisingly,
they show that the well-trained networks enable scale-consistent predictions over long videos,
while the accuracy is still inferior to traditional methods because of ignoring geometric information.
Building on top of recent progress in computer vision, we design a simple yet robust VO system by integrating multi-view geometry and deep learning on \textbf{D}epth and optical \textbf{F}low, namely \textbf{DF-VO}. 
In this work, 
a) we propose a method to \textit{carefully} sample high-quality correspondences from deep flows and recover accurate camera poses with a geometric module;
b) we address the scale-drift issue by aligning geometrically triangulated depths to the scale-consistent deep depths, where the dynamic scenes are taken into account.
Comprehensive ablation studies show the effectiveness of the proposed method,
and extensive evaluation results show the state-of-the-art performance of our system,
e.g., Ours (\textit{1.652\%}) v.s. ORB-SLAM (\textit{3.247\%}) in terms of translation error in KITTI Odometry benchmark.
Source code is publicly available at:  
\href{https://github.com/Huangying-Zhan/DF-VO}{\textcolor{blue}{DF-VO}}.

\end{abstract}


\section{Introduction}

\begin{figure}[t!]
        \centering
		\includegraphics[width=1\columnwidth]{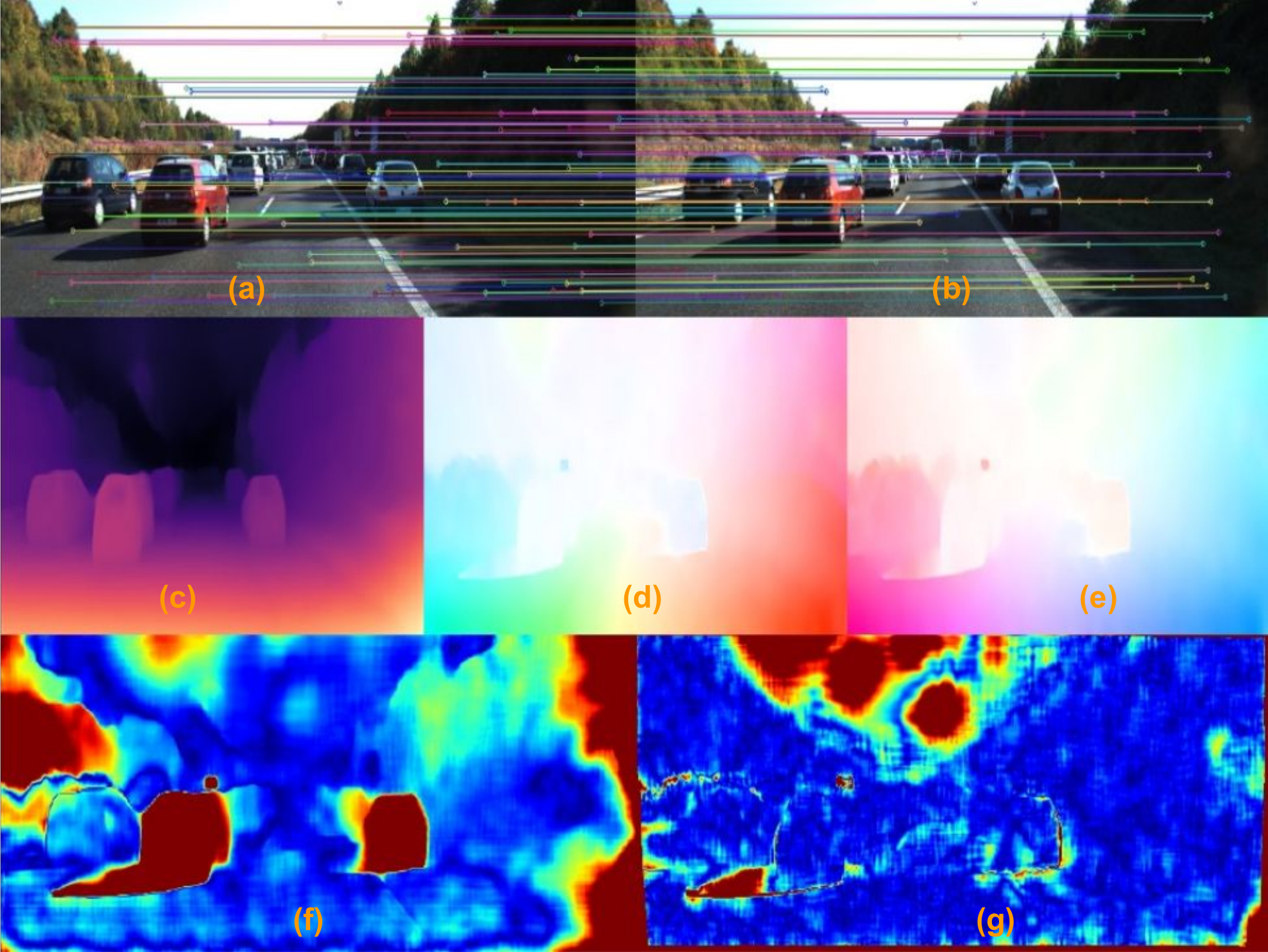}
		\caption{
		    Inputs and intermediate CNN outputs of the system. 
    		(a, b) Current and previous input images with examples of auto-selected 2D-2D matches; 
    		(c) Single view depth prediction;
    		(d, e) Forward and backward optical flow prediction;
    		(f) Flow consistency between optical flow and rigid flow;
    		(g) Forward-backward flow consistency;
    		In (f)(g), red/blue means high/low inconsistency.
		}
		\label{fig:cover_img}
\end{figure}
The ability of an autonomous robot to localize itself and know its surroundings is vital for different robotic tasks such as navigation and object manipulation.
Vision-based methods are often the preferred choice because of factors such as cost-saving, low power requirements, and useful complementary information can be provided to other sensors such as IMU, GPS, laser scanners.
We address the monocular Visual Odometry (VO) problem in this paper,
where the goal is to estimate 6DoF motions of a moving camera.

Geometry-based Visual Odometry has shown dominating performance in the last few decades,
while they are only reliable and accurate under a restrictive setup,
such as when static scenes comprising well-textured Lambertian surfaces are captured with sufficient uniform illumination enabling to establish good correspondences \cite{lowe2004sift, rublee2011orb, Bian2019gms}.
The traditional correspondence search pipeline usually detects sparse feature points firstly and then matches extracted features,
resulting in a limited number of high-quality correspondences because of the aforementioned assumptions.
The accuracy and diversity of the correspondences are of the utmost importance in solving Visual Odometry problems.
In contrast, we propose to extract accurate correspondences diversely from the dense predictions of an optical flow network using the consistency constraint between bi-directional flows.
Then the selected correspondences are fed into geometry-based trackers (Epipolar Geometry based tracker and Prospective-n-point based tracker) for accurate and robust VO estimation,
as described in \secref{sec:dfvo}.

Most monocular systems suffer from a depth-translation scale ambiguity issue, which means the predictions (structure and motion) are up-to-scale.
The scale ambiguity leads to a scale drift issue that accumulated over time.
Resolving scale-drift usually relies on keeping a scale-consistent map for map-to-frame tracking, performing an expensive global bundle adjustment for scale optimization or additional prior assumptions, like constant camera height from the known ground plane.
Recently deep learning methods have made possible end-to-end learning of structure-and-motion from unlabelled videos.
The trained single-view depth models give scale-consistent predictions with the use of stereo-based training \cite{garg2016depth, zhan2018depthVO, monodepth2} or scale-consistency constraint in monocular-based training \cite{bian2019depth}.
In this work, we propose to use the scale-consistent single-view depths as the reference to maintain a consistent scale over long videos. 
The scale-consistent depths are used in two circumstances: 
(1) scale recovery when the translation scale is missed in the Epipolar Geometry tracker; 
(2) establishing scale-consistent 3D-2D correspondences in the PnP tracker.
Besides, we propose an iterative method for robust scale recovery, which is especially effective in highly dynamic scenes by removing the extracted correspondences (i.e. outliers) on dynamic regions.

Although recent deep pose networks can learn camera motions directly from videos  \cite{wang2017deepvo, zhou2017sfmlearner, zhan2018depthVO, bian2019depth, monodepth2},
the accuracy is limited because of neglecting to incorporate geometric knowledge in inference time.
In contrast,
correspondences and scene scales are learnt in our proposed framework (\figref{fig:cover_img}).
Thus accurate camera motions are estimated using well-studied multi-view geometry in the proposed system.

To summarize, the contributions of this paper include:
\begin{itemize}
\item we propose a hybrid system, DF-VO, which leverages both deep learning and multi-view geometry for Visual Odometry.
Especially, self-supervised learning is used for training networks so expensive ground truth data is not required and it enables online finetuning.
\item we propose to sample accurate sparse correspondences from dense optical flow predictions for camera tracking,
and a bi-directional consistency based sampling method is presented.
\item we propose to use scale-consistent monocular depth predictions for maintaining a consistent scale over long video for Visual Odometry, and propose an iterative scale recovery method for better performance in dynamic scenarios.
\item the comprehensive evaluation shows that the proposed DF-VO system achieves state-of-the-art performance in standard benchmarks, and we conduct a detailed ablation study for evaluating the effect of different factors in our system.
\end{itemize}

A preliminary version of DF-VO was presented in \cite{zhan2019dfvo}.
We extend the system in the following four aspects
(1) clearer presentation and more details of the proposed system
(2) improving the system in dynamic environments with an iterative correspondence selection scheme;
(3) improving the adaptation ability in new environments by introducing an online adaptation scheme;
(4) more comprehensive experiments and ablation studies.

\section{Related Work} \label{Sec:rel_work}
\textbf{Geometry based VO: }
Camera tracking is a fundamental and well-studied problem in computer vision, with different pose estimation methods based on multiple-view geometry been established \cite{HartleyMultiView, scaramuzza2011visual}.
Early work in VO dates back to the 1980s \cite{ullman1979motion, scaramuzza2011visual}, with a successful application of it in the Mars exploration rover in 2004 \cite{matthies2007computer}, albeit with a stereo camera. Two dominant methods for geometry-based VO/SLAM are feature-based \cite{ORBSLAM2, klein2007ptam, Geiger2011viso2} and direct methods \cite{DSO,newcombe2011dtam}. The former involves explicit correspondence estimation, and the latter takes the form of an energy minimization problem based on the image colour/feature warp error, parameterized by pose and map parameters. There are also hybrid approaches that make use of the good properties of both \cite{forster2014svo,forster2016svo,engel2014lsd}.
One of the most successful and accurate full SLAM systems using a sparse (ORB) feature-based approach is ORB-SLAM2 \cite{ORBSLAM2}, along with DSO \cite{DSO}, a direct keyframe-based sparse SLAM method.
VISO2 \cite{Geiger2011viso2} on the other hand is a feature-based VO system that only tracks against a local map created by the previous two frames.
All these methods suffer from the previously mentioned issues (including scale-drift) common to monocular geometry-based systems.
Various techniques have been developed for resolving the scale drift issue.
For example,
an expensive global bundle adjustment is performed for global scale optimization based on loop-closure detection, which does not always exist \cite{ORBSLAM_2015};
or additional prior assumptions are introduced like constant camera height from the known ground plane \cite{Geiger2011viso2, zhou2019ground}.
In this work, with the aid of depth estimations from a consistent-scale deep network, scale estimation is performed with respect to the depth predictions such that a single consistent scale is maintained (\secref{sec:scale_recovery}).

\textbf{Deep learning for VO: }
For supervised learning, Agrawal \etal \cite{agrawal2015seebymoving} propose to learn good visual features from an ego-motion estimation task, in which the model is capable of relative camera pose estimation.
Wang \etal \cite{wang2017deepvo} propose a recurrent network for learning VO from videos.
Ummenhofer \etal \cite{ummenhofer2016demon} and Zhou \etal \cite{deeptam} propose to learn monocular depth estimation and VO together in an end-to-end fashion by formulating structure from motion as a supervised learning problem.
Dharmasiri \etal \cite{dharmasiri2018eng} train a depth network and extend the depth system for predicting optical flows and camera motion.
Recent works suggest that both tasks can be jointly learnt in a self-supervised manner using a photometric warp loss to replace a supervised loss based on ground truth.
SfM-Learner \cite{zhou2017sfmlearner} is the first self-supervised method for jointly learning camera motion and depth estimation.
SC-SfM-Learner \cite{bian2019depth} is a very recent work which solves the scale inconsistent issue in SfM-Learner by enforcing depth consistency.
\cite{yin2018geonet, ranjan2019cc} improve SfM-Learner by incorporating optical flow in their joint training framework for dynamics reasoning.
Some prior works solve both scale ambiguity and inconsistency issue by using stereo sequences in training \cite{li2017undeepvo, zhan2018depthVO}, which address the issue of metric scale.

The issue with the above learning-based methods is that they do not explicitly account for the multi-view geometry constraints that are introduced due to camera motion \emph{during inference}.
In order to address this, recent works propose to combine the best of learning and geometry to varying extent and degree of success. CNN-SLAM \cite{tateno2017cnnslam} fuse single view CNN depths in a direct SLAM system, and CNN-SVO \cite{loo2018cnnsvo} initialize the depth at a feature location with CNN provided depth for reducing the uncertainty in the initial map.
Yang \etal \cite{yang2018dvso} feed depth predictions into DSO \cite{DSO} as virtual stereo measurements.
Li \etal \cite{li2019posevo} refine their pose predictions via pose-graph optimisation.
In contrast to the above methods, we effectively utilize CNNs for both single-view depth prediction and correspondence estimation, on top of standard multi-view geometry to create a simple yet effective VO system.

\begin{figure*}[t]
        \centering
		\includegraphics[width=0.9\textwidth]{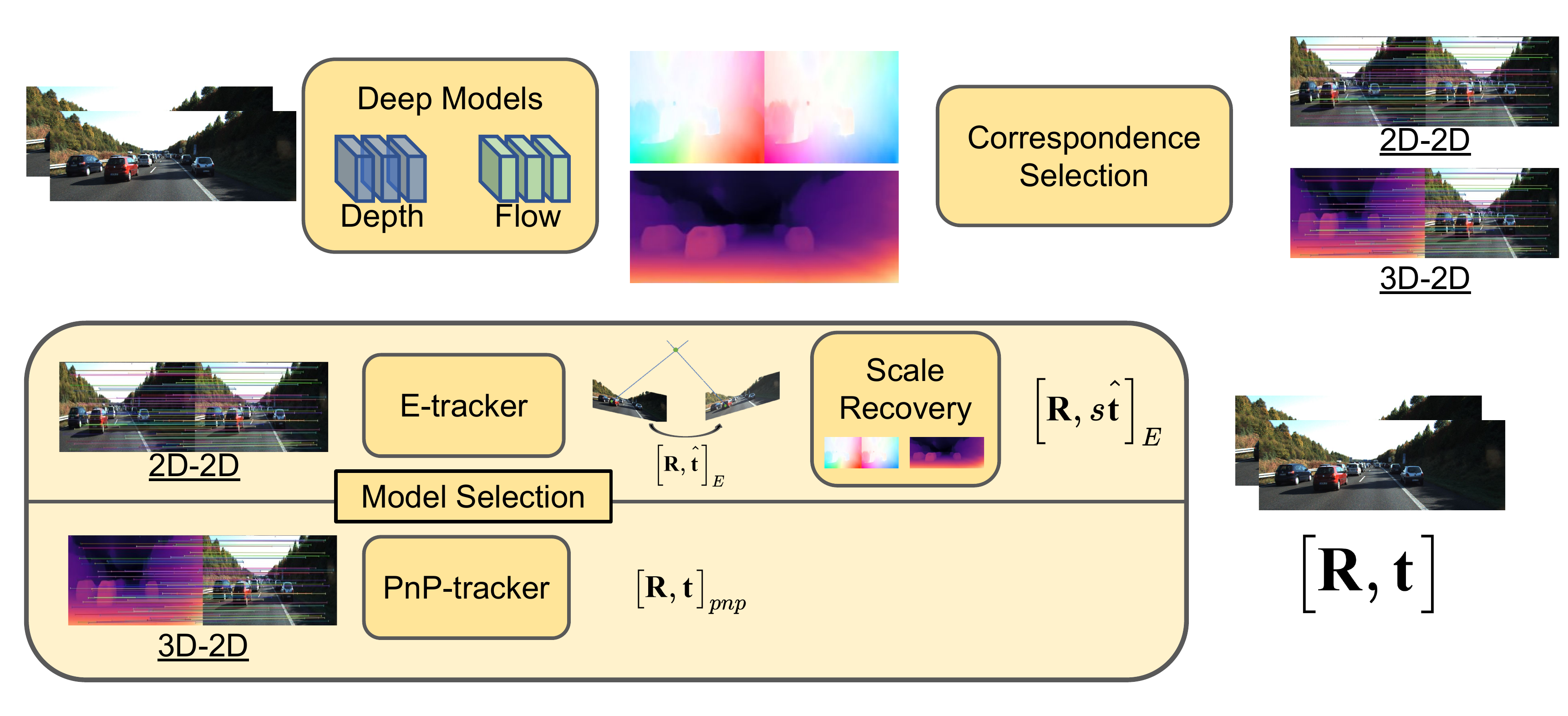}
		\caption{
		    DF-VO pipeline. For an image pair, (forward and backward) optical flows and single view depths are predicted. 
		    A forward-backward flow consistency is computed as a criterion to establish good correspondences (2D-2D; 3D-2D). 
		    We have two alternative trackers out of which one is selected by the data-driven model selection module. The first tracker (E-tracker) uses 2D-2D correspondences to estimate and decompose an essential matrix to find rotation and translation direction, which is followed by a transnational scale recovery step to estimate metric VO. 
		    The second tracker (PnP) uses single view depth estimates in conjunction with 3D-2D registration via PnP.
		}
		\label{fig:pipeline}
\end{figure*}

\section{Preliminaries}\label{sec:preliminaries}
We revisit geometry-based pose estimation methods, including Epipolar Geometry and Perspective-n-Point in this section to understand the principle and the underlying limitations of each method.
\subsection{Epipolar Geometry}\label{sec:epi_geo}
Epipolar Geometry can be employed for camera motion estimation from two images ($I_{i}, I_{j}$)
Suppose we have obtained a set of 2D-2D correspondences ($\bm{p_{i}}, \bm{p_{j}}$) from the image pair.
Epipolar constraint is employed for solving fundamental matrix, $\bm{F}$, or essential matrix, $\bm{E}$, which are related by the camera intrinsic $\bm{K}$ such that
$\bm{F} = \bm{K}^{-\bm{T}} \bm{E} \bm{K}^{-1}$.
Thus, the camera motion~$[\bm{R}, \bm{t}]$ can be recovered by decomposing $\bm{F}$ or $\bm{E}$ ~\cite{nister2003efficient,zhang1998determining,hartley1995defence,bian2019bench}.
\begin{align} \label{eqn:epipolar_constraint}
\bm{p_{j}}^{T} \bm{K}^{-\bm{T}} \bm{E} \bm{K}^{-1} \bm{p_{i}} &= 0, \text{ where } \bm{E} = [\bm{t}]_{\times}\bm{R}
\end{align}
However, the general viewpoint and general structure are assumed in such geometry guided tracking.
Problems arise with Epipolar Geometry while frames in the sequence and/or scene structure do not conform to these assumptions\cite{torr1999problem}.
\begin{itemize}
\item Motion degeneracy: motion degeneracy happens when the camera does not translate between frames, i.e. recovering $\bm{R}$ becomes unsolvable if the camera motion is a pure rotation.
\item Structure degeneracy: viewed scene structure is planar.
\end{itemize}
Solving fundamental/essential matrix becomes unstable in practice when the camera baseline is small relative to the scene size.
Moreover, translation recovered from the essential matrix is up-to-scale because of scale ambiguity.
\subsection{Perspective-n-Point}\label{sec:pnp}
Perspective-n-point (PnP) solves camera pose given known 3D-2D correspondences.
In a two-view problem, suppose we have obtained a set of 3D-2D correspondences, including the 3D points on \textit{i}-th view and the corresponding projection in \textit{j}-th view $(\bm{X_{i}}, \bm{p_{j}})$, PnP can be employed to estimate camera pose by minimizing the reprojection error,
\begin{align}
e = \sum_{x} ||\bm{K} (\bm{R}\bm{X}_{i}[\bm{x}]+\bm{t}) - \bm{p}_{j}[\bm{x}]||_{2},
\end{align}
where $[\bm{x}]$ is pixel coordinate indexing.
Solving a PnP problem requires accurate estimation of the 3D structure of the scene which can be obtained from depth sensor measurements or mature stereo reconstruction methods,
while it is a more challenging problem in the monocular case.

\section{DF-VO: Depth and Flow for Visual Odometry} \label{sec:dfvo}
\subsection{System Overview} \label{sec:overview}

A standard Visual Odometry pipeline includes feature extraction and matching to establish correspondences 
, followed by pose estimation from the correspondences.
We follow this pipeline 
and present  \textbf{DF-VO}, which is illustrated in \figref{fig:pipeline} and \algmref{alg:dfvo2}.
Two types of correspondences (2D-2D and 3D-2D) are considered in this system.
To obtain the correspondences,
(1) 
an optical flow network is trained to predict dense correspondences between images for 2D-2D correspondences establishment;
(2) a single-view depth network is used to estimate 3D structure thus 3D-2D correspondences can be established by combining the optical flow estimation.
Accurate sparse correspondences are thus selected with a carefully designed mechanism.
Two trackers used for pose estimation are named E-tracker and PnP-tracker, which employ Epipolar Geometry with a scale recovery module and Prospective-n-Point, respectively.
Note that the scale recovery module is associated with E-tracker for solving the well-known scale ambiguity and scale drift issues.
To decide a suitable tracker for each input pair,
a robust model selection method using geometric robust information criterion is used.
In order to achieve minimal training and supervision, and high-quality prediction on the deep networks, we explore a variety of training schemes on the depth network and flow network. 
Building on top of advanced deep networks and classic geometry methods, we present a simple yet effective and robust monocular Visual Odometry system.

\subsection{Deep Predictions} \label{sec:deep_pred}
In order to form 2D-2D/3D-2D correspondences from an image pair,  
specifically $(\bm{p_{i}}, \bm{p_{j}})$ or $(\bm{X_{i}}, \bm{p_{j}})$,
we propose to use an optical flow network and a single-view depth network to establish the correspondences.

\paragraph{Optical flow}
The 2D-2D correspondences are extracted from dense optical flow prediction. 
Give an image pair, $(I_{i}, I_{j})$, optical flow describes the pixel movements in $I_{i}$, which gives the correspondences of all the pixels of $I_{i}$ in $I_{j}$.
Though the state-of-the-art deep optical flow networks have shown high average accuracy,
not all the pixels share the same high accuracy. 
Therefore, we propose a correspondence selection scheme in \secref{sec:kp_sel} to pick good predictions robustly.

\paragraph{Single view depth}
In order to establish 3D-2D correspondences between two views, $(\bm{X_{i}}, \bm{p_{j}})$, we need to obtain the 3D structure of \textit{i}-th view and the correspondences between the 3D landmarks and 2D landmarks.
Traditional approaches establish the correspondences via feature matching between 3D landmarks and 2D feature points.
In this work, we use a deep depth network as our ``depth sensor'' to estimate the 3D structure on \textit{i}-th view, $\bm{X_{i}}$. 
Through the 2D-2D correspondences established by optical flows, we can directly get a set of 3D-2D correspondences and 
solve the relative camera pose by solving PnP.

Unfortunately, the current state-of-the-art single view depth estimation methods are still insufficient for recovering very accurate 3D structure (about 10\% relative error) for accurate camera pose estimation, which is shown in \tabref{table:ablation_1}. 
On the other hand, optical flow estimation is a more generic task.
The state-of-the-art deep learning methods are accurate and with good generalization ability. 
Therefore, we mainly use the 2D-2D matches for solving pose from essential matrix while the depth predictions are used for scale recovery and PnP-tracker.
As a result, PnP-tracker is used as an auxiliary tracker when E-tracker tends to fail.

\subsection{Correspondence Selection} \label{sec:kp_sel}

%
\begin{algorithm} [t] 
    \caption{DF-VO: Depth and Flow for Visual Odometry}
  \begin{algorithmic}[1] 
    \REQUIRE Depth-CNN: $M_d$; Flow-CNN: $M_f$
    \INPUT Image sequence: $[\bm{I}_1$, $\bm{I}_2$, ..., $\bm{I}_k]$
    \OUTPUT Camera poses: $[\bm{T}_1, \bm{T}_2, ..., \bm{T}_k]$
    \STATE \textbf{Initialization} $\bm{T_1} = \bm{I}$ ; $i=2$
    \WHILE{$i \leq k$}
        \STATE Get CNN predictions: $\bm{D}_i$,  $\bm{F}^{i}_{i-1}$, \text{and } $\bm{F}^{i-1}_{i}$
        \STATE Compute forward-backward flow inconsistency from ($\bm{F}^{i}_{i-1}, \bm{F}^{i-1}_{i}$).
        \STATE Correspondence selection: form matches $(\bm{P}_i, \bm{P}_{i-1})$ from the filtered flows based on flow inconsistency
        \STATE Model selection: estimate $\bm{E}$ and $\bm{H}$  from $(\bm{P}_i, \bm{P}_{i-1})$ and compute GRIC scores for the trackers
        \IF{E-Tracker}
            \STATE Recover $[\bm{R}, \uvec{t}]$ from the estimated Essential matrix 
            \STATE Triangulate $(\bm{P}_i, \bm{P}_{i-1})$ to get $\bm{D}'_{i}$
            \STATE Scale recovery to estimate $s$
            \STATE $\bm{T}^{i-1}_{i} = [\bm{R}, s\uvec{t}]$
        \ELSIF{PnP-Tracker}
            \STATE Form 3D-2D correspondences from $(\bm{D}_i, \bm{P}_i, \bm{P}_{i-1})$
            \STATE Estimate $[\bm{R}, \bm{t}]$ using PnP
            \STATE $\bm{T}^{i-1}_{i} = [\bm{R}, \bm{t}]$
        \ENDIF
        \STATE $\bm{T}_i \leftarrow \bm{T}_{i-1} \bm{T}^{i-1}_{i}$ 
    \ENDWHILE
  \end{algorithmic} \label{alg:dfvo2}
\end{algorithm}

Most deep learning-based optical flow models predict dense optical flows, i.e. every pixel is associated with a predicted flow vector. 
There can be a lot of matches formed by the optical flows, in which some of them are very accurate. 
It is time-consuming if all matches are taken into consideration in solving a VO problem since only sparse matches are required to solve the problem in theory.
The vanilla way is to sample the optical flows randomly/uniformly from the dense predictions.

\begin{figure}[t!]
            \centering
            \includegraphics[width=1\columnwidth]{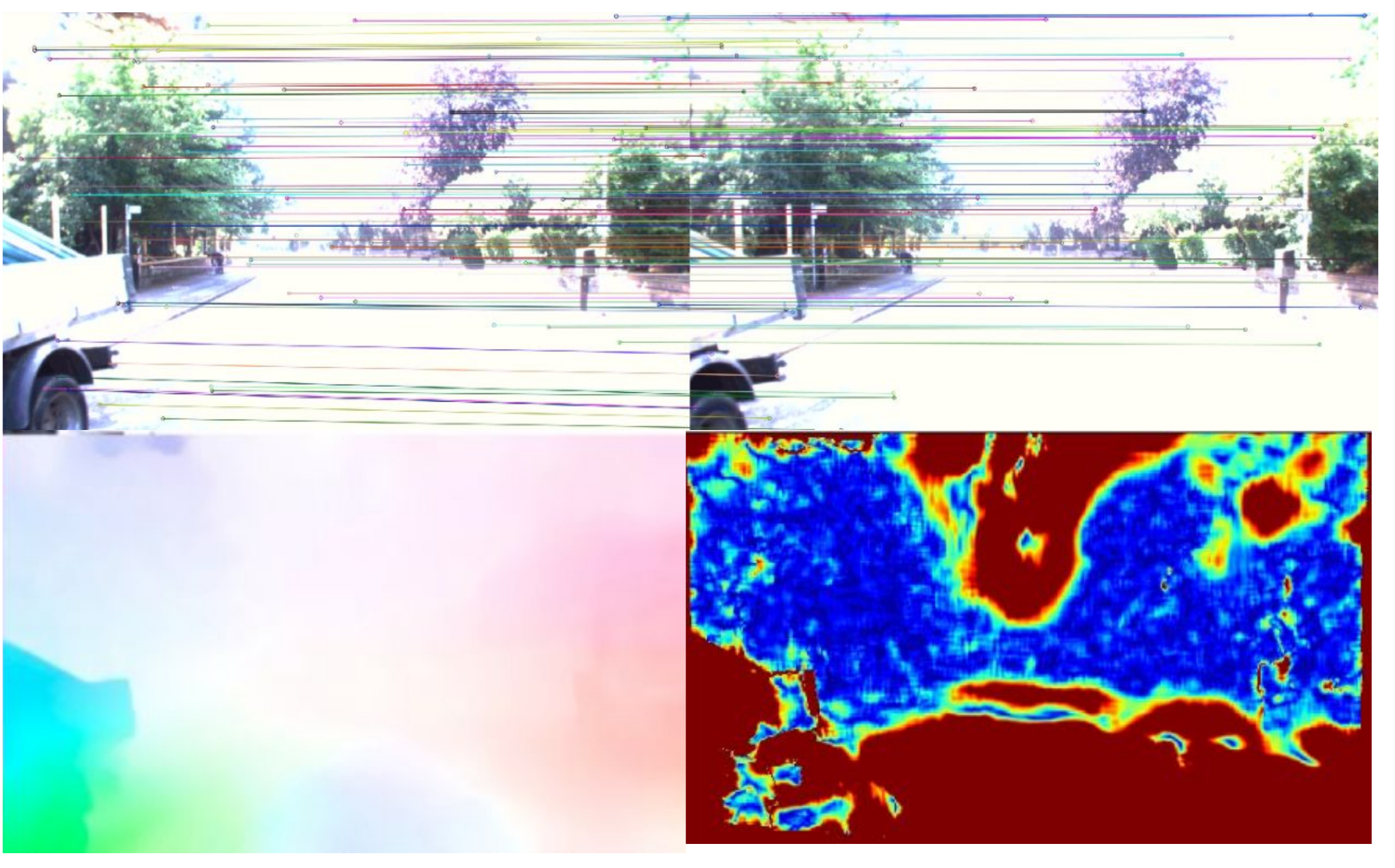}\\
\caption{ (Top) Filtered 2D correspondences established by the optical flow prediction; 
(Bottom left) Optical flow prediction; 
(Bottom right) Bidirectional flow consistency (high consistency is shown in blue) shows that sufficient correspondences can be established in the overexposure case. 
}
\label{fig:overexposure}
\end{figure}

However, we have observed that not all the flow predictions share the same high accuracy. 
Some regions in the images have worse optical flow predictions, 
for instance, 
out-of-view regions where no correspondences can be found in the other view; 
dynamic object regions where occlusion is usually associated with.
In order to filter out the outliers and pick good optical flows, we propose a correspondence selection scheme based on a bi-directional flow consistency, see example in \figref{fig:overexposure}.

\paragraph{Flow consistency}
Given an image pair, $(I_i, I_{j})$, both forward and backward optical flows, $\bm{F}^{j}_{i}$ and $\bm{F}^{i}_{j}$, are predicted by the flow network.
Thus we compute forward-backward flow consistency as a measure to choose good 2D-2D correspondences.
The flow consistency is computed by,
\begin{align}
    \bm{C} = - \bm{F}^j_{i} - w \left (\bm{F}^i_{j}, p_f(\bm{F}^{j}_{i})\right),
\end{align}
The warping process at a pixel $\bm{x}$ is described as
\begin{align}
    w\left(\bm{F}^i_{j}[\bm{x}], p_f(\bm{F}^{j}_{i}[\bm{x}])\right) 
    = \bm{F}^{i}_{j}[\bm{x} + \bm{F}^j_{i}[\bm{x}]].
\end{align}
As $\bm{x} + \bm{F}^i_{j}[\bm{x}]$ does not necessarily locate on the regular grid, the resulted flow is interpolated from the flow vectors in the 4 corners\cite{jaderberg2015stn}.
We use the flow consistency to select correspondences with higher accuracy and 
the hypothesis we made is that \textit{the optical flows with better consistency tend to have higher accuracy}, which is proved with an experiment in \secref{sec:ablation}.

\paragraph{Best-\textit{N} selection}
After computing forward-backward flow consistency, we choose optical flows with the least inconsistency $\bm{F'}$ to form the best-\textit{N} 2D-2D matches, $(\bm{P}_i, \bm{P}_{j})$ \cite{zhan2019dfvo}, where \textit{N} equals to 2000 in most experiments.
This correspondence selection scheme is able to reject a lot of inaccurate flows. As shown in \cite{zhan2019dfvo}, DF-VO with this correspondence selection scheme has already outperformed existing VO/SLAM baselines.
However, there are still some potential issues regarding the scheme.
\begin{itemize}
    \item Model under-fitting: if the chosen best-\textit{N} matches do not have enough location diversity, the pose model estimated can be an under-fitting model.
    \item Structure degeneracy: if all the chosen matches locate on a planar region, structure degeneracy happens and leads to the failure of estimating essential matrix\cite{torr1999problem}.
\end{itemize}

\paragraph{Local best-\textit{K} selection}
On top of the Best-\textit{N} selection, we want to increase the location diversity of the matches. 
We divide the image regions into \textit{M} ($M = 10\times10$) regions and choose best-\textit{K} matches from each region.
However, there might be cases that have severe inaccurate flow predictions (e.g. margin regions where usually are out-of-view) and the flow predictions should not be used.
Therefore, we first filter the flows such that only flows with inconsistency less than a threshold
can be picked.
As a result, The final correspondences $(\bm{P_i}, \bm{P_j})$ formed from $\bm{F'}$ are a union of best-\textit{K} matches in each region. 
The value \textit{K} in \textit{j}-th region is defined as
$K_j = \text{min}(N/M, Q_j)$,
where $Q_j$ is the number of valid flows after thresholding.
Since the correspondence quality is vital, we further check the number of valid correspondences and the number of regions with valid correspondences to determine if sufficient good correspondences are used.
If insufficient correspondences are found, which rarely happens (mostly when the image quality is very poor such as extreme under/over-exposure), we use a constant motion model instead of the E/PnP-tracker.

The advantages of performing local best-\textit{K} selection are two-fold,
(1) increasing location diversity as described;
(2) speeding up correspondence selection process since part of flows are rejected in the first place and sorting flow inconsistency is performed in a local image region instead of the whole image region.

Comparing to traditional feature-based methods, which only use salient feature points for matching and tracking, any pixel in the dense optical flow can be a candidate for tracking.
Moreover, traditional features usually gather visual information from local regions while CNN gathers more visual information (larger receptive field) and higher-level contextual information, which gives more accurate and robust correspondences.

After selecting good 2D-2D correspondences, the essential matrix can be solved using Epipolar Geometry as described in \secref{sec:epi_geo}. 
Then, the camera motion, consisting of rotation $\bm{R}$ and translation $\uvec{t}$, can be decomposed from the essential matrix.
However, the recovered motion is up-to-scale. 
Specifically, the translation is a unit vector representing the translation direction only.
In order to recover and maintain a consistent scale over the monocular footage, a consistent scale recovery process is required.

\subsection{Scale Recovery} \label{sec:scale_recovery}
In traditional monocular VO pipeline, the per-frame scale is recovered by aligning triangulated 3D landmarks with existing 3D landmarks which accumulates errors.

\paragraph{Simple alignment}
In this work, we use the predicted depths $\bm{D_i}$ to inform 3D structure as a reference for scale recovery.
After recovering $[\bm{R}, \uvec{t}]$ from solving essential matrix, triangulation is performed for $(\bm{P_i}, \bm{P_{j}})$ to recover up-to-scale depths $\bm{D'_i}$. 
A scaling factor, $s$,  can be estimated by aligning the triangulated depth map $\bm{D'_i}$ with the CNN depth map $\bm{D_i}$.
An important advantage of using depth CNN is that we can get rid of the scale drift issue because of the following reasons.
\begin{itemize}
    \item Depth CNN predicts per-frame 3D structures, which are scale consistent. We show that we can train scale consistent depth networks (\secref{sec:train_depth_pose}).
    \item Scale drift is introduced by an accumulated error in creating new 3D landmarks. We do not create new 3D landmarks but recover scale w.r.t. a single network.
\end{itemize}

\paragraph{Iterative alignment}
\begin{algorithm} [t] 
    \caption{Iterative Scale Recovery}
  \begin{algorithmic}[1] 
    \INPUT $[\bm{R}, \uvec{t}]$, $\bm{F'}$, $\bm{D}_i$, $s_{t-1}$
    \STATE \textbf{Initialization} $s=s_{t-1}$
    \WHILE{\text{s has not converged}}
        \STATE Pose hypothesis: $\bm{T} = [\bm{R}, s\uvec{t}]$
        \STATE Compute rigid flow $\bm{F}_{rigid}$ from $\bm{T}$ and $\bm{D}_i$
        \STATE Compute flow inconsistency: $\bm{F}_{diff}~\leftarrow~||\bm{F'} - \bm{F}_{rigid}||_2$
        \STATE Select depth-flow pairs $(\bm{D}_{i}, \bm{P}_1, \bm{P}_{2})_{sel}$ with $\bm{F}_{diff}~<~\delta_{rigid}$
        \STATE Estimate new pose, $[\bm{R}, \uvec{t}]$ from $(\bm{P}_1, \bm{P}_{2})_{sel}$
        \STATE Triangulate $(\bm{P}_1, \bm{P}_{2})_{sel}$ to get $\bm{D}'_{i}$
        \STATE Estimate scaling factor, $s_{new}$, by comparing $(\bm{D}_{i, sel}, \bm{D}'_i)$
        \STATE $s \leftarrow s_{new}$
    \ENDWHILE
  \end{algorithmic} \label{alg:iter_scale}
\end{algorithm}

Aligning 3D landmarks triangulated on selected optical flow matches with CNN depths is simple and sufficient to recover accurate scale in general cases.
However, in a highly dynamic environment, the selected optical flows can be lying on dynamic regions, which is problematic for depth alignment.
Moreover, similar to optical flow predictions, not all the predicted depths are highly accurate. 
The pixels with high forward-backward flow consistency are not guaranteed to have high depth accuracy.
Therefore, we here propose an iterative scheme, \algmref{alg:iter_scale}.

The key is to select depths and filtered optical flows (\secref{sec:kp_sel}) that are consistent with each other. 
Given that the filtered optical flows generally establish good correspondences, a pixel with depth being consistent with the optical flow means that 
(1) the pixel belongs to a static region in the environment; 
(2) the depth is likely to be accurate.
However, the depth and optical flow are related by a camera pose for a static scene. 
Since the camera pose $[\bm{R}, \uvec{t}]$ is up-to-scale and does not share the same scale with the depth prediction, we, therefore, propose an iterative approach to select depth-flow pair (\algmref{alg:iter_scale}).
We first initialize a relative pose $\bm{T}$ with a pose $\bm{T}_{0}$. Then the rigid flow is computed using the current relative pose by,
\begin{align}
  \bm{F}_{rigid} =\bm{K}\bm{T} \bm{K}^{-1}(\bm{x} \bm{D}_{i} ) - \bm{x}
\end{align}
where $\bm{x}$ belongs to pixel coordinates of the selected optical flow.
The consistency between the filtered optical flow $\bm{F'}$ and the rigid flow is then measured by 
$||\bm{F'} - \bm{F}_{rigid}||_2$.
Only depth-flow pairs with small optical-rigid flow inconsistency are selected as new matches. 
Thus, we update $\bm{T}$ with the new scaled pose and iterate the process until reaching the stopping condition (convergence or meet \textit{n}-iterations).
The scale initialization for the first image pair is set as zero while the scale at time-\textit{(t-1)} is used as the scale initialization at time-\textit{(t)}.

\subsection{Model Selection} \label{sec:model_sel}

We have presented a camera tracking method integrating Epipolar Geometry with deep predictions.
However, as mentioned in \secref{sec:epi_geo}, there are some known issues with Epipolar Geometry, i.e. motion degeneracy and unstable solution when the motion is small. 
Since we have both 3D-2D and 2D-2D correspondences available, we can instead solve a PnP problem using the correspondences obtained in \secref{sec:kp_sel} when Epipolar Geometry tends to fail. 
In this section, we show that we can select a suitable tracker/model by two possible ways.

\paragraph{Flow magnitude}
We measure the magnitude of the flow predictions and solve essential matrix only when the average flow magnitude is large enough.
It avoids small camera motions which usually come with small optical flows\cite{zhan2019dfvo}. 
However, this na\"ive approach is associated with some issues.
(1) It does not resolve motion degeneracy (pure rotation), which also causes large optical flows. 
(2) It does not take outliers into account, e.g. dynamic objects which cause optical flows even the camera is stationary. 
Therefore, we adopt a more robust measure for model selection.

\paragraph{Geometric Robust Information Criterion}
Torr\etal \cite{torr1999problem} discuss the degeneracy cases (motion and structure) and their influence on geometry guided camera motion estimation. 
Two robust strategies for tackling such degeneracies are proposed. 
(1) A statistical model selection test, named Geometric Robust Information Criterion (GRIC), is used to identify cases when degeneracies occur;
(2) multiple motion models are used to overcome the degeneracies.
In this work, we follow the first approach to identify when E-Tracker tends to fail and switch to PnP-Tracker.
\cite{torr1999problem} estimates both Fundamental $\bm{F}$ and Homography matrix $\bm{H}$ and choose the model with lower GRIC score. 
The model that explains the data best, i.e. lower GRIC score, is indicated as most likely.

GRIC calculates a score function for each tracker (Fundamental / Homography) considering the following factors.
\begin{itemize}
    \item number of matches, $n$
    \item residuals of the matches, $e_i$
    \item standard deviation of the measurement error, $\sigma$
    \item data dimension, $r$ (4 for two views)
    \item number of motion model parameters, $k$ ($5$ for $\bm{E}$, $7$ for $\bm{F}$, 8 for $\bm{H}$)
    \item dimension of the structure, $d$ (3 for $\bm{F}$, 2 for $\bm{H}$)
\end{itemize}
\begin{align}
    \text{GRIC} = \sum \rho(e^2_i)+\lambda_1 d n + \lambda_2 k
\end{align}
where $\rho(e^2_i)$ is a robust function of the residuals:
\begin{align}
    \rho(e^2) = \text{min}\left( \frac{e^2}{\sigma^2}, \lambda_3 (r-d) \right).
\end{align}
The value of the parameters are 
$\lambda_1 = \text{log}4$,
$\lambda_2 = \text{log}4n$,
$\lambda_3 = 2$.
Different from \cite{torr1999problem}, since we have both 3D-2D and 2D-2D correspondences, we can choose PnP-Tracker instead of Homography-Tracker when E-Tracker tends to fail.

\paragraph{Cheirality condition}
In addition to the two methods introduced above, we check for cheirality condition as well. 
There are 4 possible solutions for $[\bm{R}, \uvec{t}]$ by decomposing $\bm{E}$. 
To find the correct unique solution, cheirality condition, i.e. the triangulated 3D points must be in front of both cameras, is checked to remove the other solutions. 
We further use the number of points satisfying cheirality condition as a reference to determine if the solution is stable.

Therefore, we choose PnP-Tracker when $\text{GRIC}_E$ is higher than $\text{GRIC}_H$ or cheirality check condition is not fulfilled. 
Otherwise, E-Tracker is employed for solving frame-to-frame camera motion.
To robustify the system, we wrap the trackers in RANSAC loops. 

\subsection{Jointly learning of depths and pose} \label{sec:train_depth_pose}
Various depth training frameworks can be employed depending on the availability of data (monocular/stereo sequences, depth sensor measurements).
The most trivial way is using a supervised training framework \cite{eigen2014depth, liu2015depth, liu2016depth, laina2016deeper, kendall2017uncertainties, nekrasov2019multitask, fu2018dorn}, but ground truth depths are not always available for any scenario. 
Some recent works suggest that jointly learning single-view depths and camera motion in a self-supervised manner is feasible using monocular sequences \cite{zhou2017sfmlearner, yin2018geonet,monodepth2, bian2019depth}, or stereo sequences \cite{garg2016depth, godard2017monodepth, zhan2018depthVO, monodepth2}.
Instead of using ground truth supervisions, the main supervision signal in the self-supervised framework is photometric consistency across multiple-views.

In this work, we mainly follow \cite{monodepth2} for training depth models using monocular and stereo sequences.
The depth network is based on the encoder-decoder architecture with skip connections\cite{ronneberger2015u}.
The pose network consists of a ResNet18 feature extractor which takes an image pair as input (concatenated as a 6-channel input) and predicts 6-DoF relative pose. 
We refer readers to \cite{monodepth2} for more network architecture details.

\subsubsection{Training overview} 
In this work, we jointly train the depth network and the pose network by minimizing the mean of the following \textit{per-pixel} objective function over the whole image.
The \textit{per-pixel} loss is
\begin{align}
    L = &\min_{j}L_{pe}(\bm{I}_{i}, \bm{I}^{i}_{j})  + 
        \lambda_{ds} L_{ds}(\bm{D}_i, \bm{I}_i)  + \nonumber \\ 
        &\min_{j} \lambda_{dc} L_{dc}(\bm{D}_{i}, \bm{D}^{i}_{j}), \label{eqn:total_Loss}
\end{align}
where $L_{pe}$ is photometric loss; 
$L_{ds}$ is depth smoothness loss; 
$ L_{dc}$ is depth consistency loss;
and $[\lambda_{ds}, \lambda_{dc}]$ are loss weightings.

\subsubsection{Photometric loss}
$L_{pe}$ is the photometric error by computing the difference between the reference image $\bm{I}_i$ and the synthesized view $\bm{I}^{i}_{j}$ warped from the source image $\bm{I}_j$, where $j \in [i-n, i+n, s]$. 
$[i-n, i+n]$ are neighbouring views of $I_i$ while $s$ is stereo pair if stereo sequences are used in training. 
As proposed in \cite{monodepth2}, instead of averaging the photometric errors between the reference pixel and the synthesized pixels from multiple views, \cite{monodepth2} only counts the photometric error between the reference pixel and the synthesized pixel with the minimum error.
The rationale is to overcome the issues related to out-of-view pixels and occlusions. 
\begin{align}
    L_{pe}(\bm{I}_{i}, \bm{I}^{i}_{j}) &= 
    \frac{\alpha}{2}\left(1-\text{SSIM}(\bm{I}_i, \bm{I}^i_j)\right) + 
    (1-\alpha)|\bm{I}_i-\bm{I}^i_j| \\ 
    \bm{I}^i_j &= w \left(\bm{I}_j, p_{re}(\bm{K}, \bm{D}_i, \bm{T}^j_i)\right) \label{eqn:photo_warp}, 
\end{align}
where SSIM \cite{wang2004image} is a robust measurement for image similarity and $\alpha = 0.85$ balances the SSIM error and the simple color intensity error.
$w(\bm{I}, \bm{p})$ is a differentiable warping function \cite{jaderberg2015stn} which warps image $\bm{I}$ according to the pixel locations $\bm{p}$.
$p_{re}(\bm{K}, \bm{D}_i, \bm{T}^j_i)$ establishes the pixel coordinates reprojected from view-\textit{i} to view-\textit{j}, where $\bm{K}$ is the camera intrinsics, $\bm{D}_i$ is the predicted depth map of view-\textit{i}, and $\bm{T}^j_i$ is the relative pose between the pair.
The reprojection for a pixel $\bm{x}$ from view-\textit{i} to view-{j} is represented by
\begin{align}
    p_{re}\left(\bm{K}, \bm{D}_i, \bm{T^j_i}\right) = \bm{K}\bm{T^j_i}\bm{K^{-1}}\bm{x} \bm{D}_{i}[\bm{x}]
\end{align}

\subsubsection{Depth smoothness regularization}
Following the approach in \cite{godard2017monodepth}, we encourage depth to be smooth locally so we induce an edge-aware depth smoothness term.
The depth discontinuity is penalized if colour continuity is presented in the same local region.
The smoothness regularization is formulated as 
\begin{align}
    L_{ds}(\bm{D}_i, \bm{I}_i) = 
        |\partial_x \bm{D}_{i}| e^{-|{\partial_x \bm{I}_{i}}|}+
        |\partial_y \bm{D}_{i}| e^{-|{\partial_y \bm{I}_{i}}|}, \label{eqn:depth_sm_Loss}
\end{align}
where $\partial_x(.)$ and $\partial_y(.)$ are gradients in horizontal and vertical direction respectively. 
Note that we use inverse depth regularization instead.

\subsubsection{Training without scaling issues} \label{sec:train_depth_consistency}
Similar to traditional monocular 3D reconstruction, \textbf{scale ambiguity} and \textbf{scale inconsistency} issues exist when monocular videos are used for training. 
Since the monocular training usually uses image snippets (usually 2 or 3 frames) for training, the training does not guarantee a consistent learnt scale across snippets and it creates the scale inconsistency issue\cite{bian2019depth}.

One solution to solve both scale problems is using stereo sequences during training \cite{li2017undeepvo, zhan2018depthVO, monodepth2}, the deep predictions are aligned with real-world scale and scale-consistent because of the constraint introduced by the known stereo baseline. 
Even though stereo sequences are used during training, only monocular images are required during inference for depth predictions.

Another solution to overcome the scale inconsistency issue is using temporal geometry consistency regularization proposed in \cite{zhan2019depthnormal, bian2019depth}, which constrains the depth consistency across multiple views. 
As depth predictions are consistent across different views and thus different snippets, the scale inconsistency issue is resolved.
Using the rigid scene assumption as the cameras move in space over time we want the predicted depths at view-\textit{i} to be consistent with the respective predictions at view-\textit{j}.
This is done by \textbf{correctly transforming} the scene geometry from frame-$j$ to frame-$i$ much like the image warping.
Specifically, we adopt the inverse depth consistency proposed in \cite{zhan2019depthnormal}. 
\begin{align}
    L_{dc}(\bm{D}_i, \bm{D}^i_j) = |1/\bm{D}_{i} - 1/\bm{D}^i_j|
\end{align}
Inspired by \cite{monodepth2}, we use minimum error in multi-view pairs to avoid occlusions and out-of-view scenes instead of averaging the depth consistency error over all source views.

\subsection{Learning of optical flows} \label{sec:train_flow}
Many deep learning-based methods have been proposed for estimating optical flow \cite{dosovitskiy2015flownet, ilg2017flownet2, hui18liteflownet, sun2018pwc, meister2018unflow}.
In this work, we choose LiteFlowNet\cite{hui18liteflownet} as our backbone network for optical flow prediction since LiteFlowNet is fast, lightweight, and accurate. 
LiteFlowNet consists of a two-stream network for feature extraction and a cascaded network for flow inference and regularization.
We refer readers to \cite{hui18liteflownet} for more details.
LiteFlowNet shows good generalization ability.
LiteFlowNet trained on a synthetic dataset (Scene Flow\cite{dosovitskiy2015flownet}) can generalize well in real-world scenarios, though sometimes artifacts present in some regions.

In this work, we mainly use the model trained from Scene Flow.
However, we also show that a self-supervised finetuning can be performed to help the model better adapt to unseen environments and remove the artifacts.
Two finetuning schemes are tested and compared, including offline finetuning and online finetuning (\secref{sec:ablation_flow}).
Similar to the self-supervised training of the depth network, the optical flow network is trained by minimizing the mean of the following \textit{per-pixel} loss function over the whole image.
\begin{align}\label{eqn:flow_Loss}
    L = &\min_{j}L_{pe}(\bm{I}_{i}, \bm{I}^{i}_{j})  + 
        \lambda_{fs} L_{fs}(||\bm{F}^j_i||_2, \bm{I}_i) \nonumber \\ 
        & + \lambda_{fc} L_{fc}\left(\left|-\bm{F}^j_i - w\left(\bm{F}^i_j, p_f(\bm{F}^j_i) \right) \right| \right)  \\
    \bm{I}^i_j &= w\left(\bm{I}_j, p_f(\bm{F}^j_i)\right), 
\end{align}
Different from \eqnref{eqn:photo_warp}, $p_f(.)$ establish the correspondences between view-\textit{i} and view-\textit{j} via the flow field instead of using reprojection defined in \eqnref{eqn:photo_warp}.
For a pixel $\bm{x}$ on view-\textit{i}, the corresponding pixel position , $p_f(\bm{F}^{j}_{i}[\bm{x}])$, on view-\textit{j} is
$\bm{x} + \bm{F}^j_{i}[\bm{x}]$.

We also regularize the optical flow to be smooth using an edge-aware flow smoothness loss $L_{fs}(.)$ similar to the depth smoothness loss defined in \eqnref{eqn:depth_sm_Loss}.
Similar to Meister \etal \cite{meister2018unflow}, we estimate both forward and backward optical flow and constrain the bidirectional predictions to be consistent with the loss $L_{fc}$.

\section{Implementation and Benchmarking} \label{sec:exp_eval}
\subsection{Dataset}
We train and test our method in popular benchmarking datasets, KITTI
\cite{Geiger2012kitti, Geiger2013kitti} and Oxford Robotcar\cite{RobotCarDatasetIJRR}, which are large scale outdoor driving datasets.
There are various splits in KITTI for several tasks, e.g. depth estimation, odometry, object tracking.
In this work, we select the following three splits to evaluate our method.

\paragraph{KITTI Odometry}
Odometry split contains 11 driving sequences with publicly available ground truth camera poses. Most of the sequences are long sequences and some with loop closing.
Following \cite{zhou2017sfmlearner}, we train our networks on sequences 00-08. The dataset contains 36,671 training pairs, [$I_i, I_{i-1}, I_{i+1}, I_{i,s}$].

 \paragraph{KITTI Tracking}
Tracking split contains 21 sequences with available ground truths.
The split is primarily used for object tracking benchmarking so there are more dynamic objects in these sequences when compared to the Odometry split, but shorter sequence length in general.
Following \cite{zhang2020vdo}, we choose 9 out of the 21 sequences with a considerable number of dynamic objects to test the robustness of our system in dynamic environments.
These sequences are challenging for most monocular VO/SLAM systems since most of the systems assume static scenarios.
\paragraph{KITTI Flow}
KITTI Flow 2012/2015 splits contain 194/200 image pairs with high-quality optical flow labels.
We use this split to evaluate the performance of the optical flow models in this work.
\paragraph{Oxford Robotcar}
To further test the generalization ability of the system, we test the proposed system on Oxford Robotcar dataset.
Following \cite{loo2018cnnsvo}, 8 sequences are selected for evaluation and the first 200 frames
\footnote{Our system can operate even without skipping the frames. The 200 frames are skipped in the evaluation for a fair comparison.}
are skipped in the evaluation due to the extremely overexposed images at the beginning of the sequences.
 
\subsection{Deep network training}
We train our networks with the PyTorch \cite{paszke2017pytorch} framework. 
All self-supervised experiments are trained using Adam optimizer \cite{kingma2014adam} for 20 epochs. 
For KITTI, images with a size of 640 $\times$ 192 are used for training.
Learning rate is set to $10^{-4}$ for the first 15 epochs and then is dropped to $10^{-5}$ for the remaining epochs.
The loss weightings are $[\lambda_{ds}, \lambda_{dc}]=[10^{-3}, 5]$ for jointly learning depths and camera motion 
while $[\lambda_{fs}, \lambda_{fc}]=[10^{-1}, 5 \times 10^{-3}]$ for optical flow experiments.

\subsection{Visual Odometry Benchmarking}
\begin{figure}[t!]
    \begin{multirow}{3}{*}
        \centering
        \begin{multicols}{2}
            \centering
            \includegraphics[width=1\columnwidth]{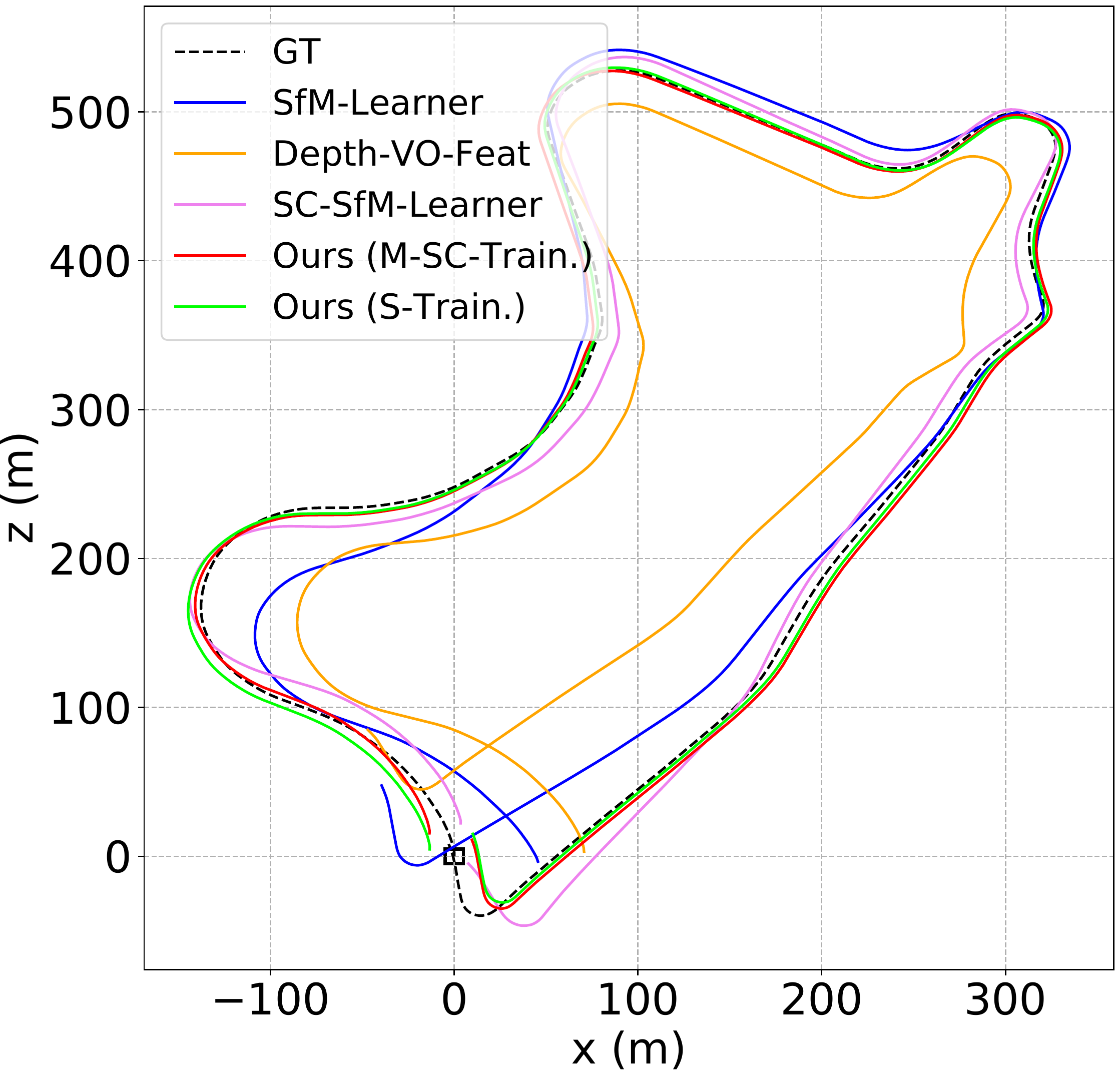}\\
            \includegraphics[width=1\columnwidth]{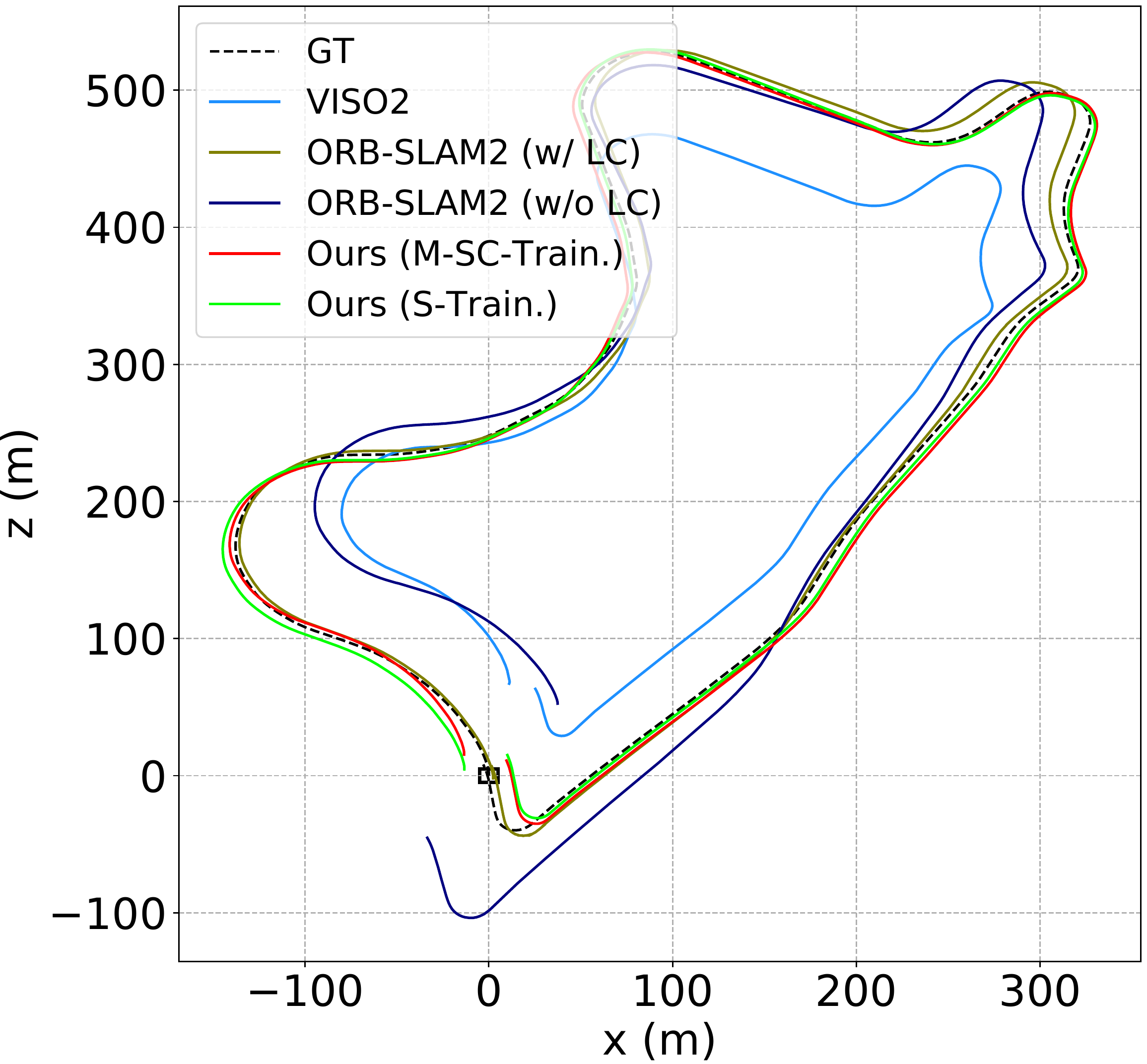}\\
        \end{multicols}
        \begin{multicols}{2}
            \centering
            \includegraphics[width=1\columnwidth]{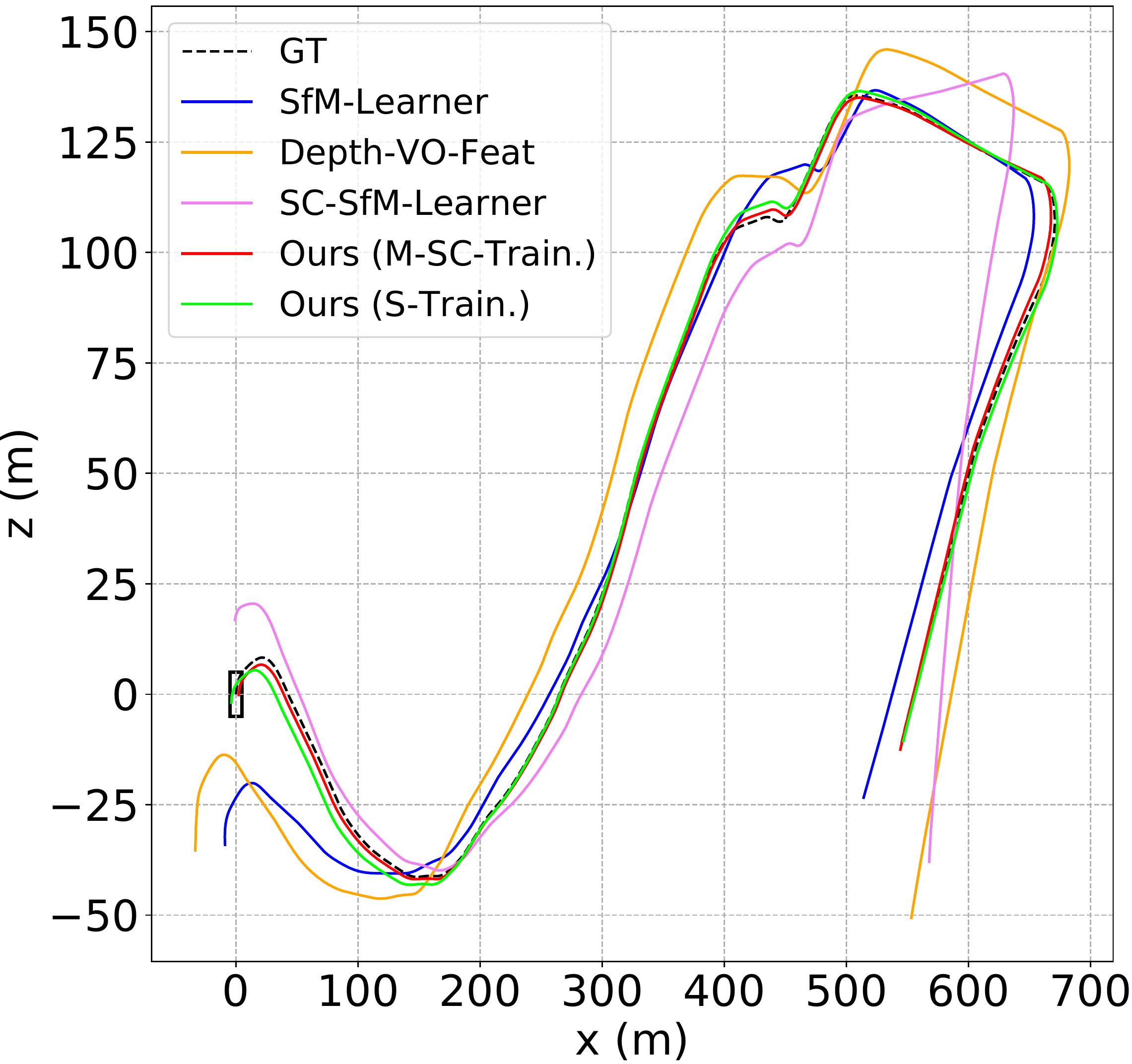}\\
            \includegraphics[width=1\columnwidth]{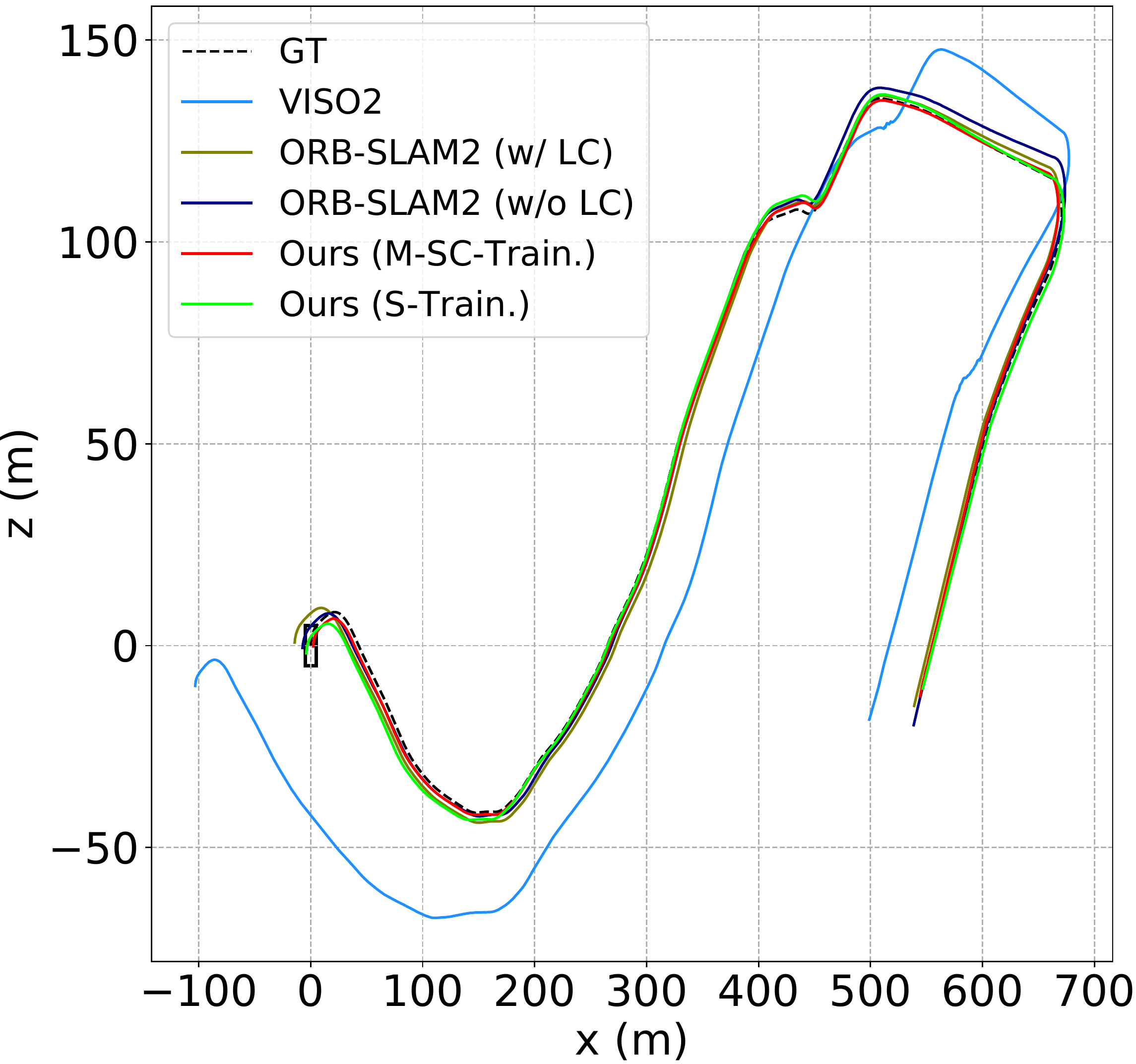}\\
        \end{multicols}
        \vspace{-20pt}
    \end{multirow}
\caption{Qualitative VO results on KITTI: (Top) Seq.09 and (Bottom) Seq.10 against deep learning-based and geometry-based methods (shown separately).
}
\label{fig:traj}
\end{figure}
\begin{figure*}[t!]
    \begin{multirow}{2}{*}
        \centering
        \begin{multicols}{4}
            \centering
            \includegraphics[width=1\columnwidth]{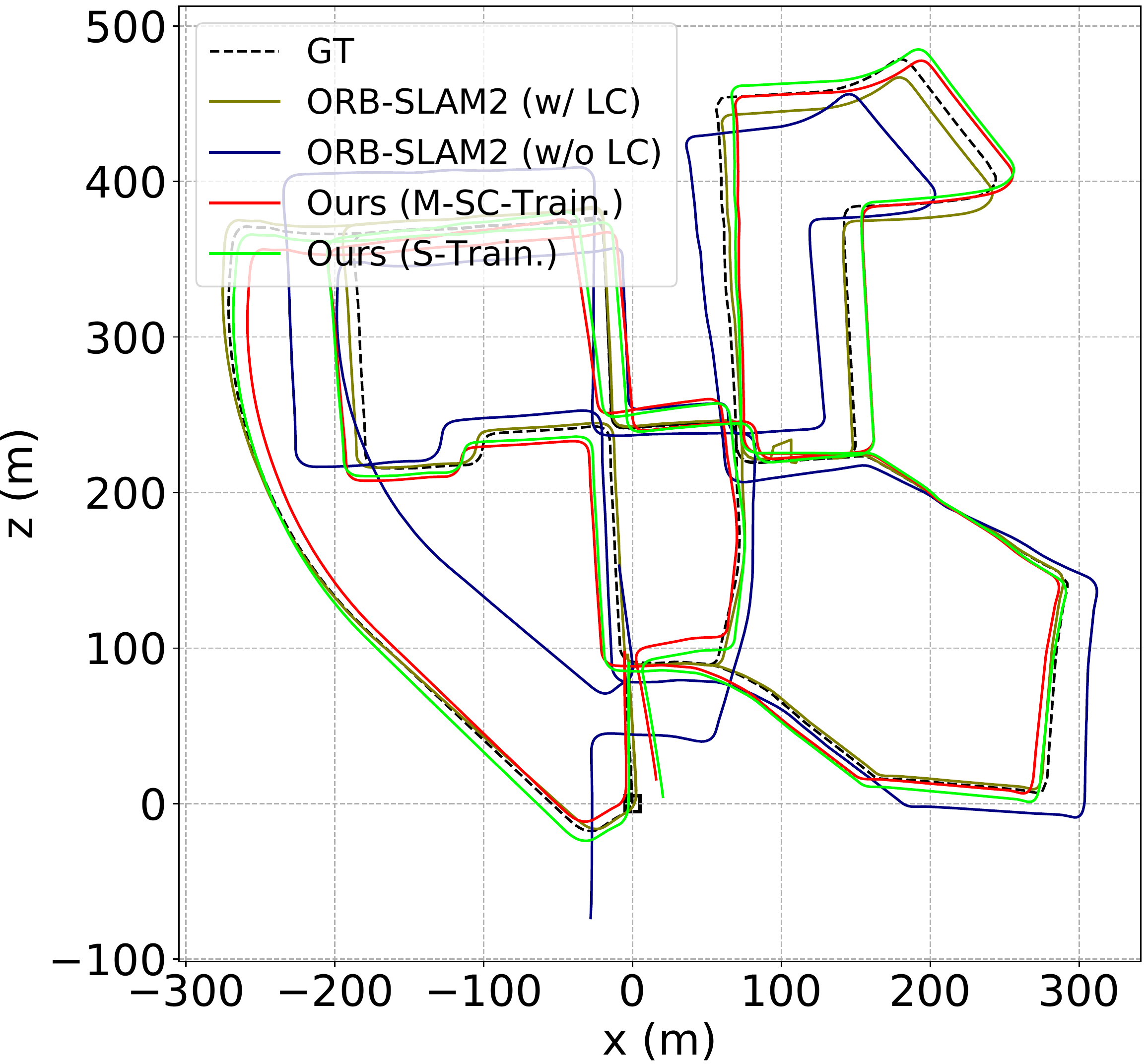}\\
            \includegraphics[width=1\columnwidth]{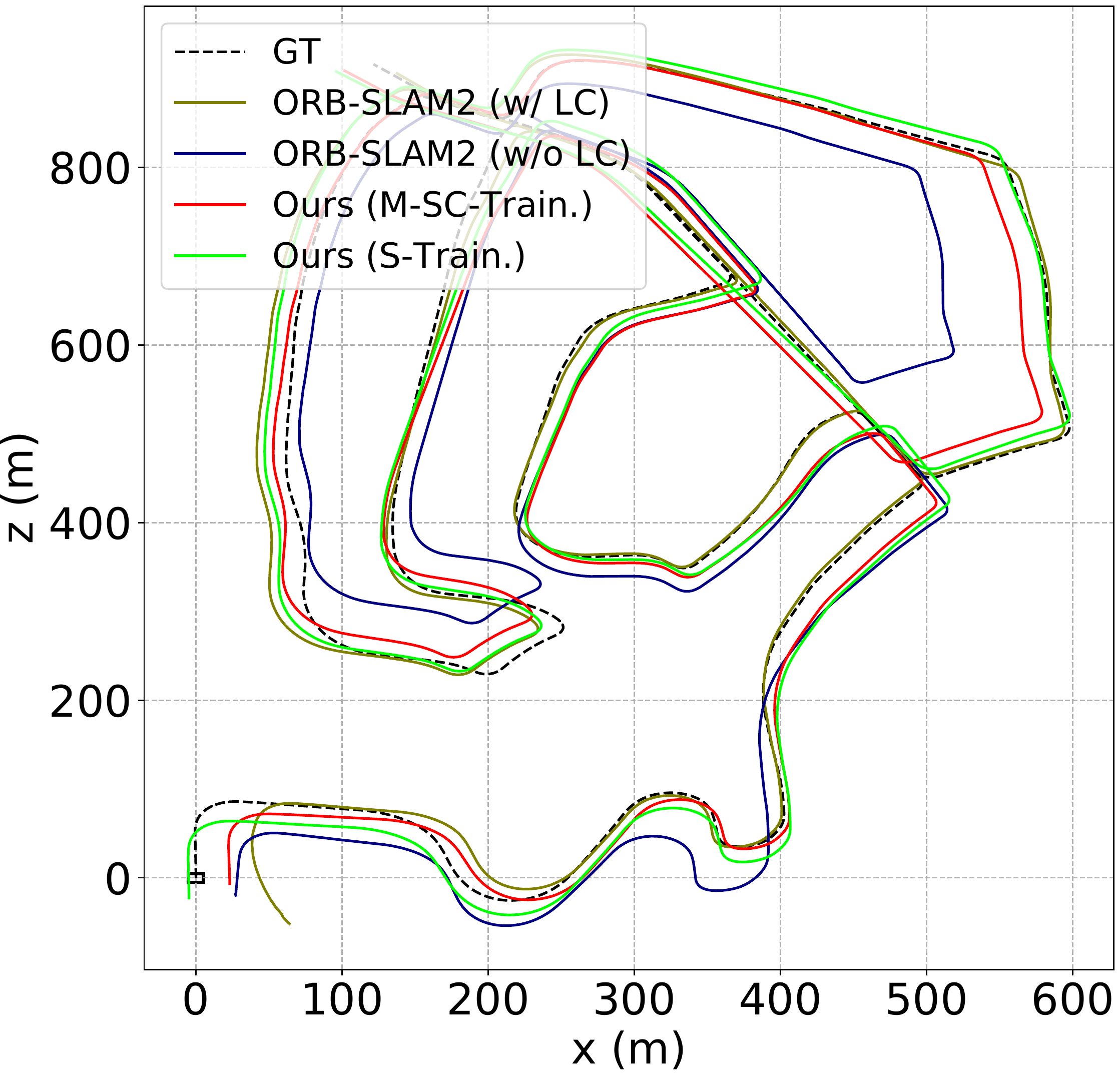}\\
            \includegraphics[width=1\columnwidth]{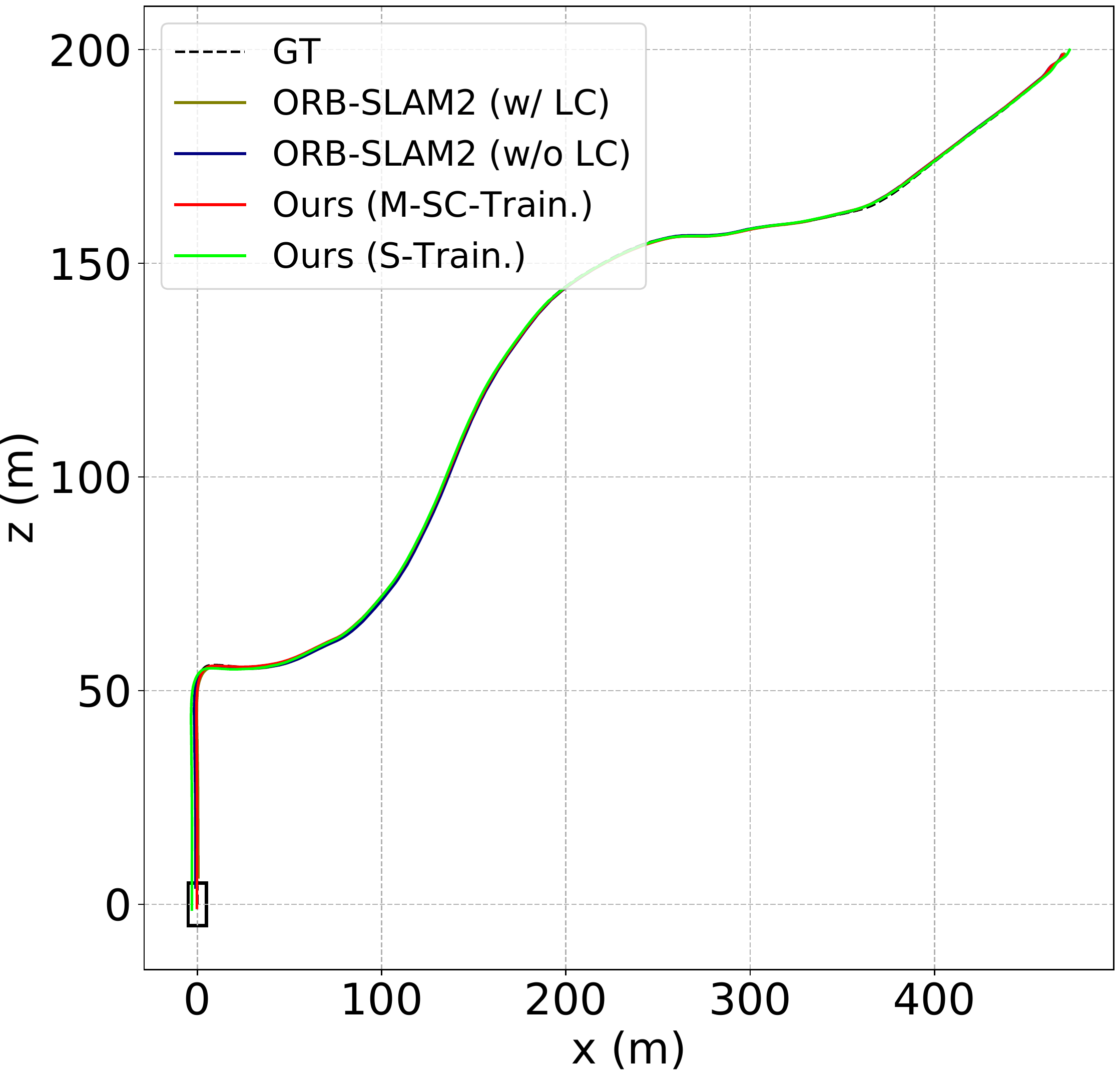}\\
            \includegraphics[width=1\columnwidth]{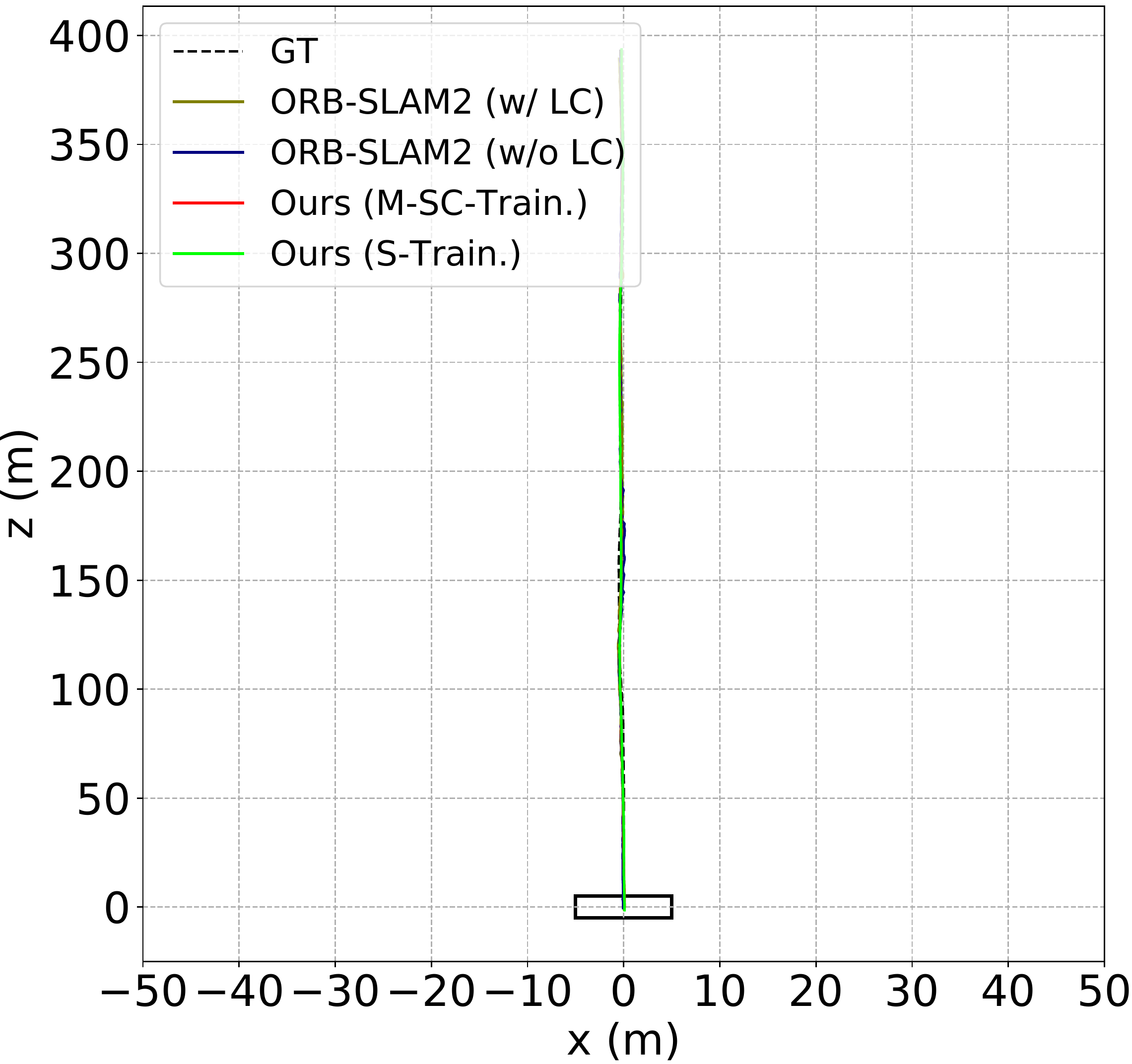}\\
        \end{multicols}
        \begin{multicols}{4}
            \centering
            \includegraphics[width=1\columnwidth]{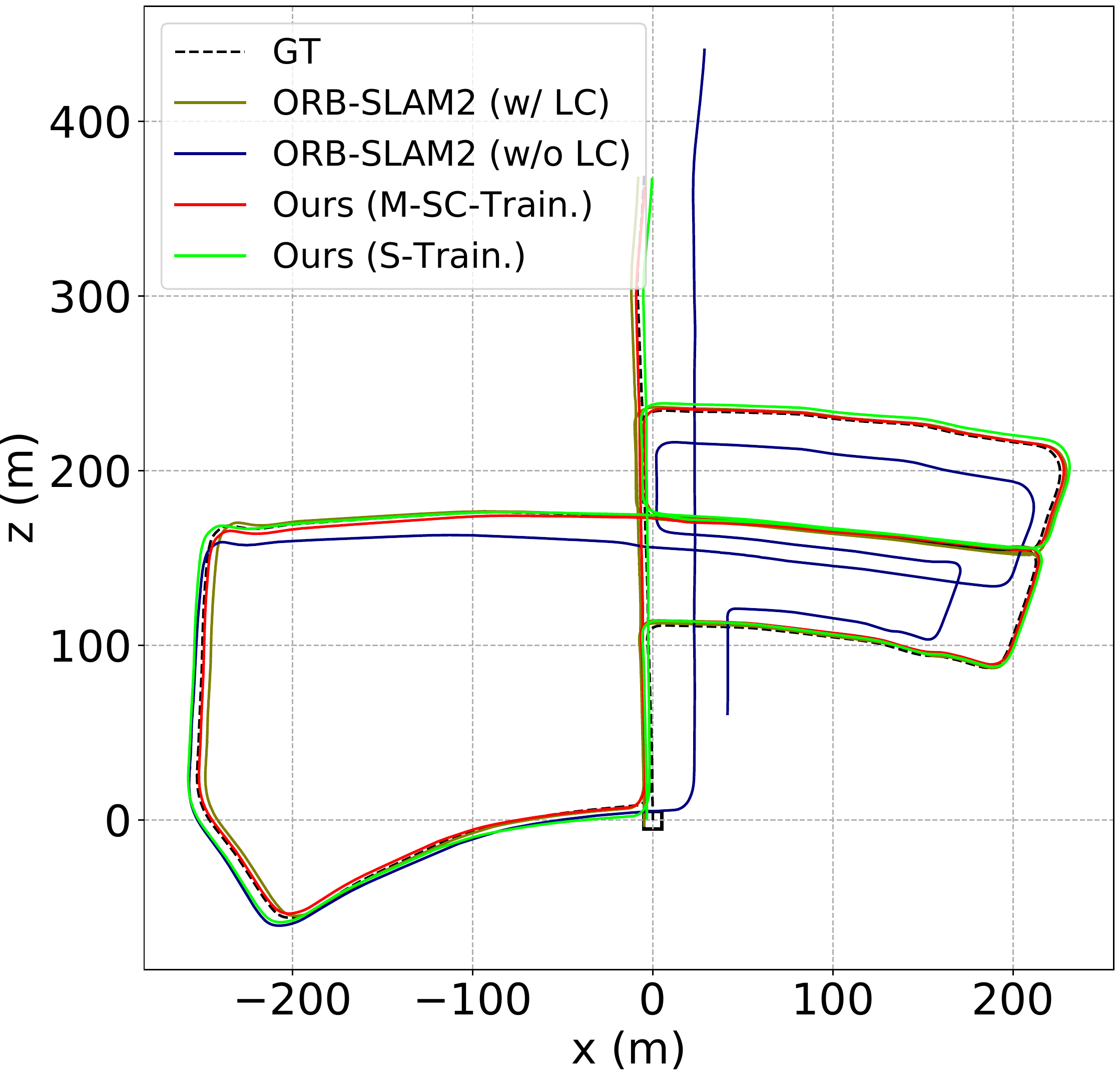}\\
            \includegraphics[width=1\columnwidth]{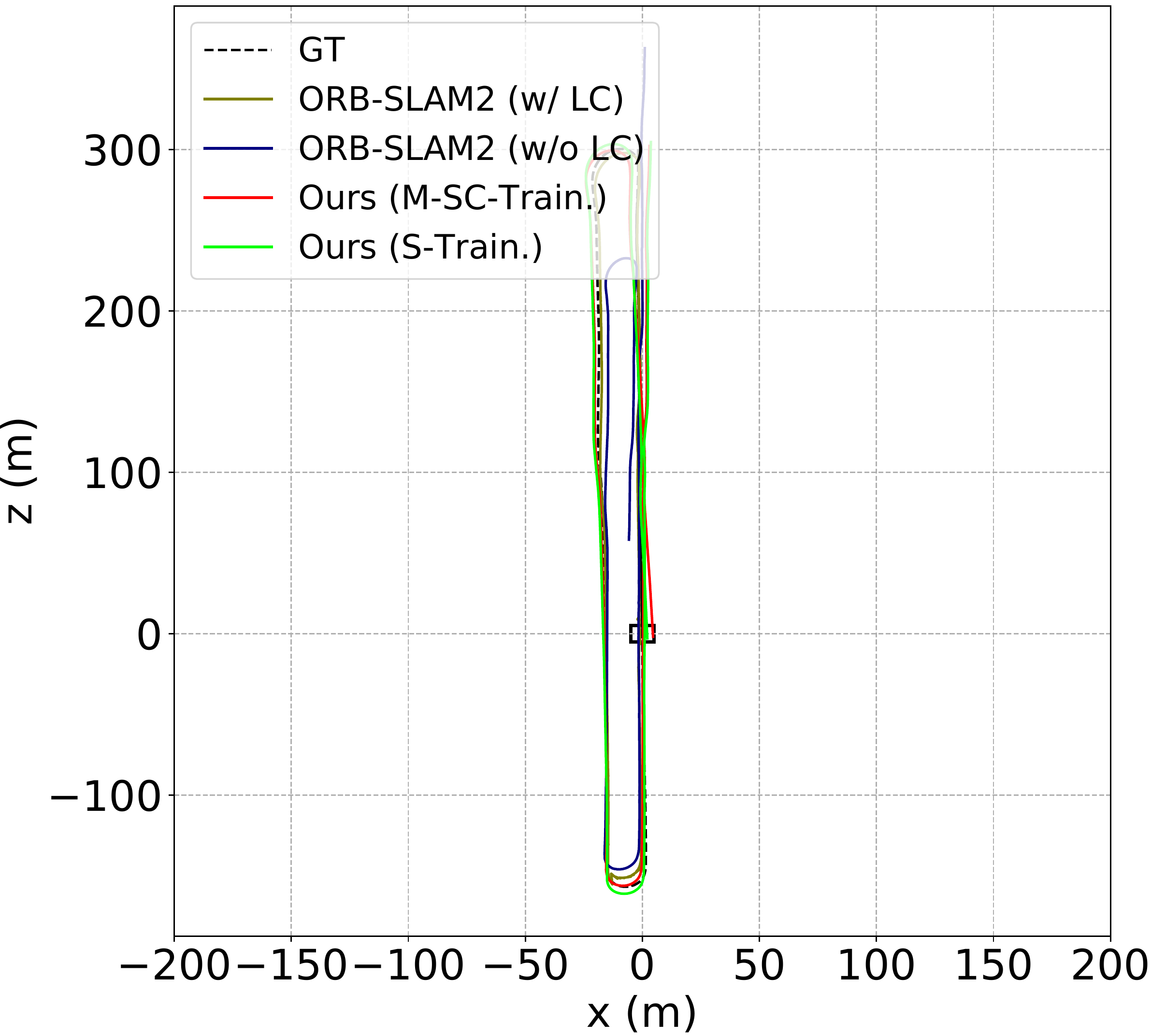}\\
            \includegraphics[width=1\columnwidth]{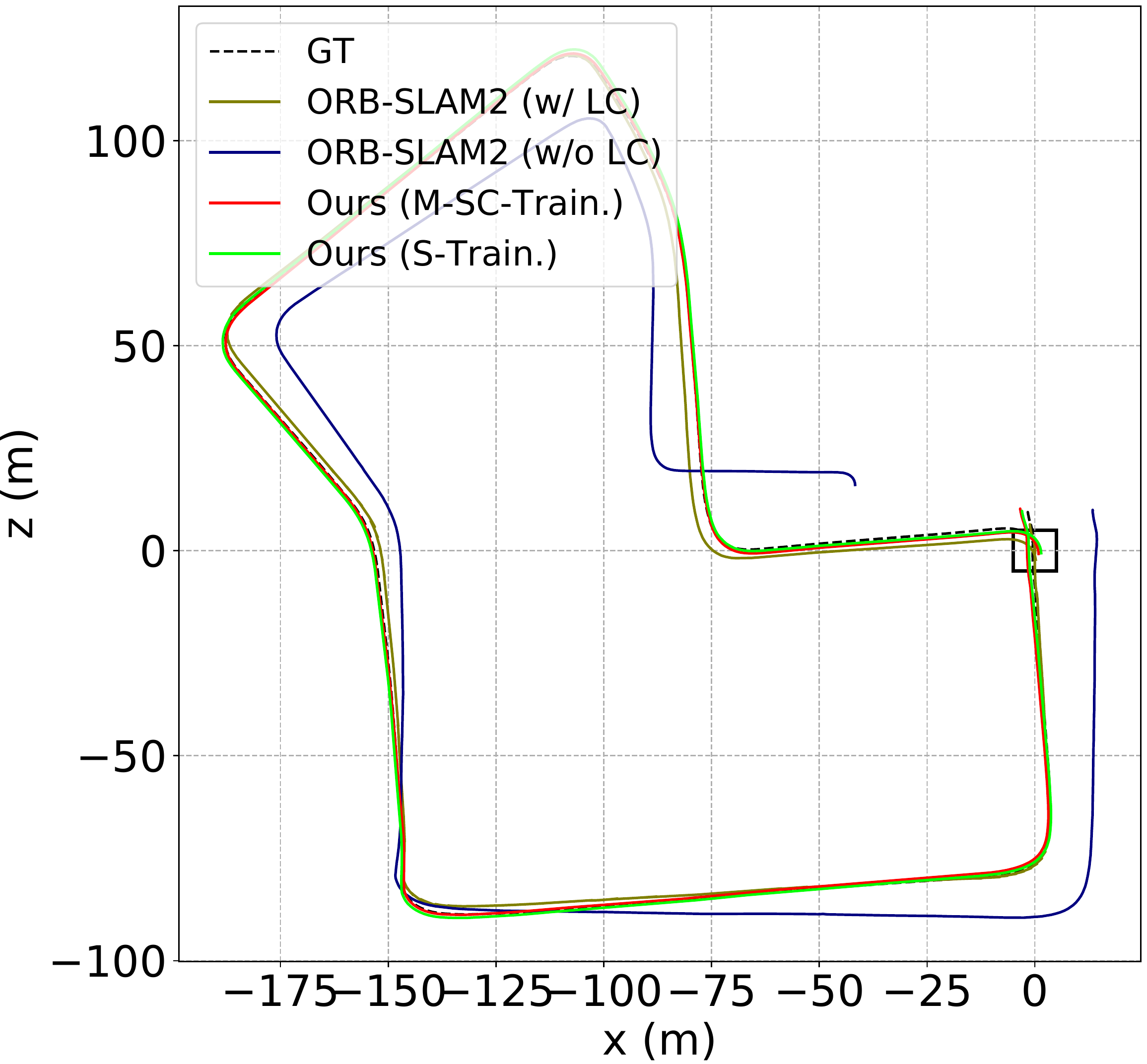}\\
            \includegraphics[width=1\columnwidth]{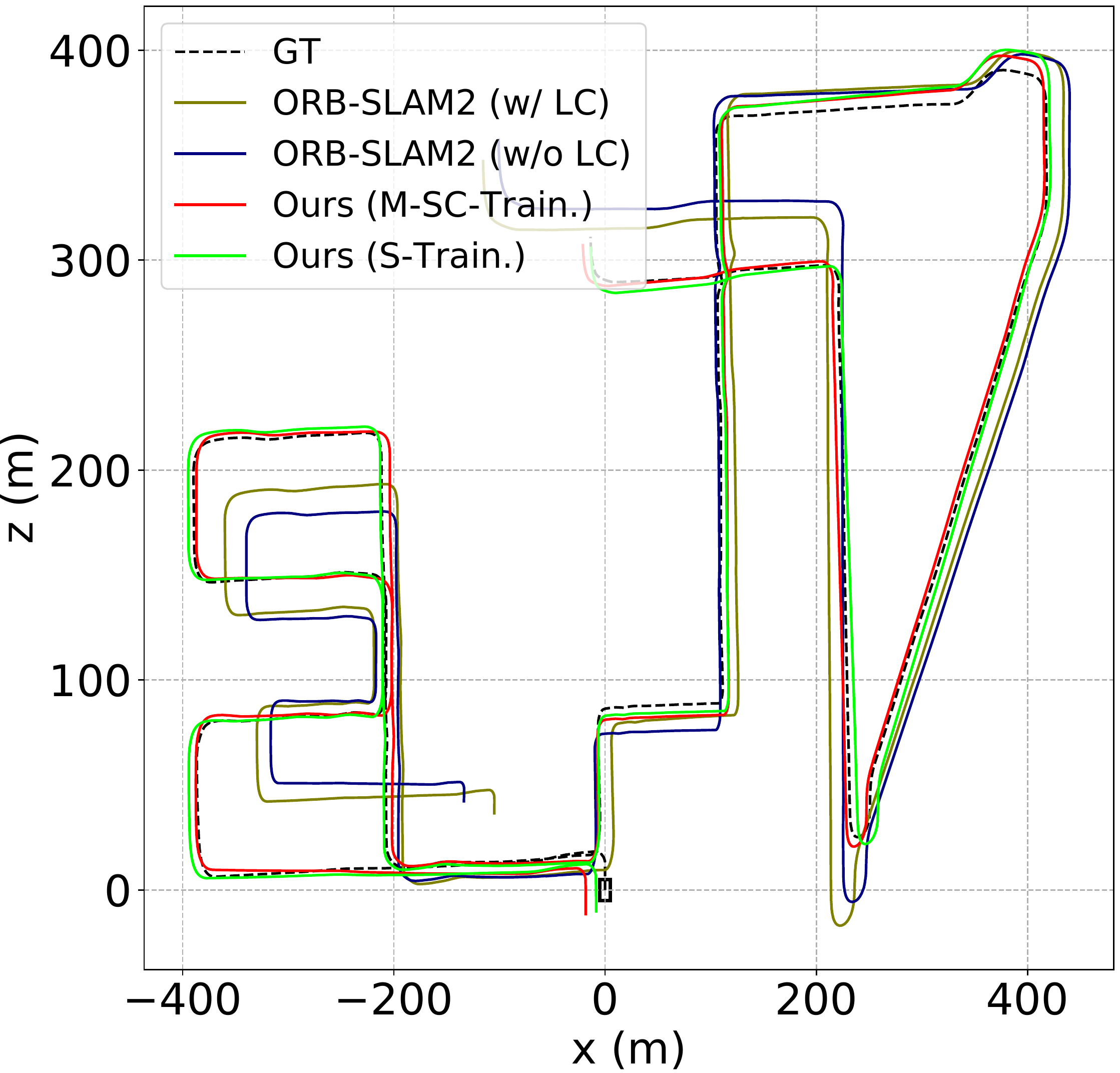}\\
        \end{multicols}
    \end{multirow}
\caption{DF-VO and ORB-SLAM2 (monocular, w/ and w/o loop-closure) trajectories in sequences 00, 02, 03, 04, 05, 06, 07 and 08 from the KITTI odometry benchmark. Note that Seq. 08 does not contains loops and ORB-SLAM2 (w/ LC) undergoes severe scale drifting while DF-VO does not.
}
\label{fig:traj2}
\end{figure*}
%

\begin{table*} [!t] 
\caption{
Quantitative result on KITTI Odometry Seq. 00-10. The best result is in bold and second best is underlined.
}
\begin{center}
\resizebox{2\columnwidth}{!}{%
\begin{tabular}{| c | c  | c | c c c c c c c c c c c | c |}
\hline
Category & Method & Metric & 00 & 01 & 02 & 03 & 04 & 05 & 06 & 07 & 08 & 09 & 10 & Avg. Err.\\
\hline\hline
%
\multirow{15}*{Deep VO} &
\multirow{5}*{\shortstack[1]{SfM-Learner \\\cite{zhou2017sfmlearner}}} & 
$t_{err}$   & 
21.32 & \textbf{22.41} & 24.10 & 12.56 & 4.32 & 
12.99 & 15.55 & 12.61 & 10.66 & 11.32 & 15.25 & 14.068 \\ 
& & $r_{err}$ & 
6.19 & 2.79 & 4.18 & 4.52 & 3.28 & 
4.66 & 5.58 & 6.31 & 3.75 & 4.07 & 4.06 & 4.660 \\ 
& & ATE & 
104.87 & \underline{109.61} & 185.43 & 8.42 & 3.10 & 
60.89 & 52.19 & 20.12 & 30.97 & 26.93 & 24.09 & 51.701 \\ 
& & RPE (m) & 
0.282 & \underline{0.660} & 0.365 & 0.077 & 0.125 & 
0.158 & 0.151 & 0.081 & 0.122 & 0.103 & 0.118 & 0.158 \\ 
& & RPE ($^\circ$) &  
0.227 & 0.133 & 0.172 & 0.158 & 0.108 & 
0.153 & 0.119 & 0.181 & 0.152 & 0.159 & 0.171 & 0.160 \\ 
\cline{2-15} 

 &
\multirow{5}*{\shortstack[1]{Depth-VO-Feat \\\cite{zhan2018depthVO}}} & 
$t_{err}$   & 
6.23 & \underline{23.78} & 6.59 & 15.76 & 3.14 & 
4.94 & 5.80 & 6.49 & 5.45 & 11.89 & 12.82 & 7.911 \\ 
& & $r_{err}$ & 
2.44 & 1.75 & 2.26 & 10.62 & 2.02 & 
2.34 & 2.06 & 3.56 & 2.39 & 3.60 & 3.41 & 3.470 \\ 
& & ATE & 
64.45 & 203.44 & 85.13 & 21.34 & 3.12 & 
22.15 & 14.31 & 15.35 & 29.53 & 52.12 & 24.70 & 33.220 \\ 
& & RPE (m) & 
0.084 & \textbf{0.547} & 0.087 & 0.168 & 0.095 & 
0.077 & 0.079 & 0.081 & 0.084 & 0.164 & 0.159 & 0.108 \\ 
& & RPE ($^\circ$) &  
0.202 & 0.133 & 0.177 & 0.308 & 0.120 & 
0.156 & 0.131 & 0.176 & 0.180 & 0.233 & 0.246 & 0.193 \\ 
\cline{2-15} 

&
\multirow{5}*{\shortstack[1]{SC-SfMLearner \\\cite{bian2019depth}}} & 
$t_{err}$   & 
11.01 & 27.09 & 6.74 & 9.22 & 4.22 & 
6.70 & 5.36 & 8.29 & 8.11 & 7.64 & 10.74 & 7.803 \\ 
& & $r_{err}$ & 
3.39 & 1.31 & 1.96 & 4.93 & 2.01 & 
2.38 & 1.65 & 4.53 & 2.61 & 2.19 & 4.58 & 3.023 \\ 
& & ATE & 
93.04 & \textbf{85.90} & 70.37 & 10.21 & 2.97 & 
40.56 & 12.56 & 21.01 & 56.15 & 15.02 & 20.19 & 34.208 \\ 
& & RPE (m) & 
0.139 & 0.888 & 0.092 & 0.059 & 0.073 & 
0.070 & 0.069 & 0.075 & 0.085 & 0.095 & 0.105 & 0.086 \\ 
& & RPE ($^\circ$) &  
0.129 & 0.075 & 0.087 & 0.068 & \underline{0.055} & 
0.069 & 0.066 & 0.074 & 0.074 & 0.102 & 0.107 & 0.083 \\ 
\hline \hline


\multirow{13}*{VO with Optim.} &
\multirow{1}*{DSO \cite{DSO}} & 
ATE & 
113.18 & / & 116.81 & 1.39 & \textbf{0.42} & 
47.46 & 55.62 & 16.72 & 111.08 & 52.23 & 11.09 & 52.600 \\ 
\cline{2-15} 

\multirow{8}*{Full SLAM /}  & 
\multirow{5}*{\shortstack[1]{ORB-SLAM2 (w/o LC) \\\cite{ORBSLAM2}}} & 
$t_{err}$   & 
11.43 & 107.57 & 10.34 & \underline{0.97} & \underline{1.30} & 
9.04 & 14.56 & 9.77 & 11.46 & 9.30 & 2.57 & 8.074 \\ 
& & $r_{err}$ & 
\underline{0.58} & 0.89 & \textbf{0.26} & \textbf{0.19} & \underline{0.27} & 
0.26 & \underline{0.26} & 0.36 & \textbf{0.28} & 0.26 & \underline{0.32} & \underline{0.304} \\ 
& & ATE & 
40.65 & 502.20 & 47.82 & \textbf{0.94} & 1.30 & 
29.95 & 40.82 & 16.04 & 43.09 & 38.77 & 5.42 & 26.480 \\ 
& & RPE (m) & 
0.169 & 2.970 & 0.172 & 0.031 & 0.078 & 
0.140 & 0.237 & 0.105 & 0.192 & 0.128 & 0.045 & 0.130 \\ 
& & RPE ($^\circ$) &  
0.079 & 0.098 & \underline{0.072} & \underline{0.055} & 0.079 & 
\underline{0.058} & 0.055 & 0.047 & 0.061 & 0.061 & \underline{0.065} & \underline{0.063} \\ 
\cline{2-15} 

& 
\multirow{5}*{\shortstack[1]{ORB-SLAM2 (w/ LC) \\\cite{ORBSLAM2}}} & 
$t_{err}$   & 
2.35 & 109.10 & 3.32 & \textbf{0.91} & 1.56 & 
1.84 & 4.99 & 1.91 & 9.41 & 2.88 & 3.30 & 3.247 \\ 
& & $r_{err}$ & 
\textbf{0.35} & \textbf{0.45} & \underline{0.31} & \textbf{0.19} & \underline{0.27} & 
\textbf{0.20} & \textbf{0.23} & \textbf{0.28} & \underline{0.30} & 0.25 & \textbf{0.30} & \textbf{0.268} \\ 
& & ATE & 
\textbf{6.03} & 508.34 & \textbf{14.76} & 1.02 & 1.57 & 
\underline{4.04} & 11.16 & 2.19 & 38.85 & 8.39 & 6.63 & 9.464 \\ 
& & RPE (m) & 
0.206 & 3.042 & 0.221 & 0.038 & 0.081 & 
0.294 & 0.734 & 0.510 & 0.162 & 0.343 & 0.047 & 0.264 \\ 
& & RPE ($^\circ$) &  
0.090 & 0.087 & 0.079 & \underline{0.055} & 0.076 & 
0.059 & 0.053 & 0.050 & 0.065 & 0.063 & 0.066 & 0.066 \\ 
\hline \hline

\multirow{15}*{VO} &
\multirow{5}*{\shortstack[1]{VISO2 \\\cite{Geiger2011viso2}}} & 
$t_{err}$   & 
10.53 & 61.36 & 18.71 & 30.21 & 34.05 & 
13.16 & 17.69 & 10.80 & 13.85 & 18.06 & 26.10 & 19.316 \\ 
& & $r_{err}$ & 
2.73 & 7.68 & 1.19 & 2.21 & 1.78 & 
3.65 & 1.93 & 4.67 & 2.52 & 1.25 & 3.26 & 2.519 \\ 
& & ATE & 
79.24 & 494.60 & 70.13 & 52.36 & 38.33 & 
66.75 & 40.72 & 18.32 & 61.49 & 52.62 & 57.25 & 53.721 \\ 
& & RPE (m) & 
0.221 & 1.413 & 0.318 & 0.226 & 0.496 & 
0.213 & 0.343 & 0.191 & 0.234 & 0.284 & 0.442 & 0.297 \\ 
& & RPE ($^\circ$) &  
0.141 & 0.432 & 0.108 & 0.157 & 0.103 & 
0.131 & 0.118 & 0.176 & 0.128 & 0.125 & 0.154 & 0.134 \\ 
\cline{2-15} 

& 
\multirow{5}*{\shortstack[1]{\textbf{Ours} \\\textbf{(Mono-SC Train.)}}} & 
$t_{err}$   & 
\underline{2.33} & 39.46 & \underline{3.24} & 2.21 & 1.43 & 
\textbf{1.09} & \textbf{1.15} & \textbf{0.63} & \underline{2.18} & \underline{2.40} & \textbf{1.82} & \underline{1.848} \\ 
& & $r_{err}$ & 
0.63 & 0.50 & 0.49 & 0.38 & 0.30 & 
\underline{0.25} & 0.39 & \underline{0.29} & 0.32 & \underline{0.24} & 0.38 & 0.367 \\ 
& & ATE & 
14.45 & 117.40 & 19.69 & \underline{1.00} & 1.39 & 
\textbf{3.61} & \textbf{3.20} & \textbf{0.98} & \underline{7.63} & \underline{8.36} & \textbf{3.13} & \underline{6.344} \\ 
& & RPE (m) & 
\underline{0.039} & 1.554 & \underline{0.057} & \underline{0.029} & \underline{0.046} & 
\underline{0.024} & \underline{0.030} & \underline{0.021} & \underline{0.041} & \underline{0.051} & \underline{0.043} & \underline{0.038} \\ 
& & RPE ($^\circ$) &  
\underline{0.056} & \textbf{0.049} & \textbf{0.045} & \textbf{0.038} & \textbf{0.029} & 
\textbf{0.035} & \textbf{0.029} & \textbf{0.030} & \underline{0.037} & \textbf{0.036} & \textbf{0.043} & \textbf{0.038} \\ 
\cline{2-15} 

& 
\multirow{5}*{\shortstack[1]{\textbf{Ours} \\\textbf{(Stereo Train.)}}} & 
$t_{err}$   & 
\textbf{2.01} & 40.02 & \textbf{2.32} & 2.22 & \textbf{0.74} & 
\underline{1.30} & \underline{1.42} & \underline{0.72} & \textbf{1.66} & \textbf{2.07} & \underline{2.06} & \textbf{1.652} \\ 
& & $r_{err}$ & 
0.61 & \underline{0.47} & 0.48 & \underline{0.30} & \textbf{0.25} & 
0.30 & 0.32 & 0.35 & 0.33 & \textbf{0.23} & 0.36 & 0.353 \\ 
& & ATE & 
\underline{12.17} & 342.71 & \underline{17.59} & 1.96 & \underline{0.70} & 
4.94 & \underline{3.73} & \underline{1.06} & \textbf{6.96} & \textbf{7.59} & \underline{4.21} & \textbf{6.091} \\ 
& & RPE (m) & 
\textbf{0.025} & 0.854 & \textbf{0.030} & \textbf{0.021} & \textbf{0.026} & 
\textbf{0.018} & \textbf{0.025} & \textbf{0.015} & \textbf{0.030} & \textbf{0.044} & \textbf{0.040} & \textbf{0.027} \\ 
& & RPE ($^\circ$) &  
\textbf{0.055} & \underline{0.052} & \textbf{0.045} & \textbf{0.038} & \textbf{0.029} & 
\textbf{0.035} & \underline{0.030} & \underline{0.031} & \textbf{0.036} & \underline{0.037} & \textbf{0.043} & \textbf{0.038} \\ 
\cline{2-15} 

\hline

%
%

\end{tabular}
}
\end{center}
\vspace{-10pt}
\label{table:kitti_benchmark}
\end{table*}

\paragraph{Evaluation Criterion}
Some common evaluation criteria are adopted for a detailed analysis.
KITTI Odometry criterion reports the average translational error $t_{err} (\%)$ and rotational errors $r_{err} (^\circ/100m)$ by evaluating possible sub-sequences of length (100, 200, ..., 800) meters.
Absolute trajectory error (ATE) measures the root-mean-square error between predicted camera poses $[x, y, z]$ and ground truth.
Relative pose error (RPE) measures frame-to-frame relative pose error.
Since most of the methods are a monocular method, which lacks a scaling factor to match with the real-world scale, we scale and align (7DoF optimization) the predictions to the ground truth associated poses during evaluation by minimizing ATE \cite{umeyama1991least}. 
Except for methods using stereo depth models (Ours(Stereo Train.), Depth-VO-Feat) and known scale prior (VISO2), which have already aligned predictions to real-world scale,
for a fair comparison, we perform 6DoF optimization w.r.t ATE instead.

\paragraph{KITTI Odometry}
We provide a detailed comparison between our VO system and some prior arts in KITTI Odometry split, which includes pure deep learning methods \cite{zhou2017sfmlearner}\footnote{SfM-Learner\cite{zhou2017sfmlearner}: the updated model in Github is evaluated}, \cite{zhan2018depthVO} \cite{bian2019depth}, and geometry-based methods including DSO\cite{DSO}\footnote{result taken from \cite{loo2018cnnsvo}}, 
VISO2\cite{Geiger2011viso2}, and ORB-SLAM2~\cite{mur2015orbslam}~(w/ and w/o loop-closure).
ORB-SLAM2 occasionally suffers from tracking failure or unsuccessful initialization. 
We run ORB-SLAM2 three times and report the one with the least trajectory error.
The quantitative and qualitative results are shown in \tabref{table:kitti_benchmark}, \figref{fig:traj}, and \figref{fig:traj2}.
Seq.01 is not included while computing average error since a sub-sequence of Seq.01 does not contain trackable close features and most methods fail in the sub-sequence.

Ours (Mono-SC Train.) uses a depth model trained with monocular videos and inverse depth consistency for ensuring scale-consistency.
Ours (Stereo Train) uses a depth model trained with stereo videos. 
Note that even stereo sequences are used during training, monocular sequences are used in testing.
Therefore, Ours (Stereo Train) is still a monocular VO system.
We show that our methods outperform pure deep learning methods, which rely on a PoseCNN for camera motion estimation, by a large margin in all metrics.
For KITTI Odometry criterion, ORB-SLAM2 shows less rotation drift $r_{err}$ but higher translation drift $t_{err}$ due to scale drift issue, which is also showed in \figref{fig:traj}.
The drifting issue sometimes can be resolved by loop closing with expensive global bundle adjustment but the issue exists when there is no loop closing detected.
Different from other methods, we use a single depth network as our ``reference map". 
The translation scales are recovered w.r.t to the scale-consistent depth predictions. 
As a result, we mitigate the scale drift issue in most monocular VO/SLAM systems and show less translation drift over long sequences.
More importantly, our method shows a consistently smaller relative error, both translation and rotation, which allows our system to be a robust module for frame-to-frame tracking.

\paragraph{KITTI Tracking}
To show the robustness of our system in \textbf{dynamic environments}, we compare our system with ORB-SLAM2 in KITTI Tracking dataset individually. 
The results are shown in \tabref{table:kitti_tracking}.
However, since the Tracking split contains relatively shorter sequences when compared to the Odometry split, KITTI Odometry criterion is not a suitable measurement to evaluate the performance. 
Therefore, we report frame-to-frame RPE (translation) for Tracking split as a reference.
Note that sequence (2011/10/03-47) is the most difficult sequence among the 9 sequences due to its highly dynamic environment in a highway.
ORB-SLAM2 is well known for its superior ability in removing outliers but its performance still downgraded significantly in this sequence while our method performs robustly.

\begin{table}[t!] 
\caption{
Visual odometry evaluation in Oxford Robotcar Dataset. 
Absolute Trajectory Error (metre) is used as the evaluation criterion.
}

\begin{center}
\resizebox{1\columnwidth}{!}{%
\begin{tabular}{| c  | c | c | c | c | c |}
\hline
\multirow{2}*{Sequence} & 
SVO & 
CNN-SVO & 
DSO &
ORB-SLAM (w/o LC) &
\multirow{2}*{\textbf{Ours}} 
\\
 & 
 \cite{forster2016svo} & 
 \cite{loo2018cnnsvo}& 
 \cite{DSO}&
\cite{ORBSLAM_2015} &
\\

\hline\hline

2014-05-06-
& \multirow{2}*{X}
& \multirow{2}*{8.66}
& \multirow{2}*{4.71}
& \multirow{2}*{10.66}
& \multirow{2}*{\textbf{4.16}}
\\
12-54-54 & & & & & \\

2014-05-06-
& \multirow{2}*{X}
& \multirow{2}*{9.19}
& \multirow{2}*{X}
& \multirow{2}*{X}
& \multirow{2}*{\textbf{3.46}}
\\
13-09-52 & & & & & \\

2014-05-06-
& \multirow{2}*{X}
& \multirow{2}*{10.19}
& \multirow{2}*{X}
& \multirow{2}*{X}
& \multirow{2}*{\textbf{4.55}}
\\
13-14-58 & & & & & \\

2014-05-06-
& \multirow{2}*{X}
& \multirow{2}*{8.26}
& \multirow{2}*{X}
& \multirow{2}*{X}
& \multirow{2}*{\textbf{4.58}}
\\
13-17-51 & & & & & \\

2014-05-14-
& \multirow{2}*{X}
& \multirow{2}*{13.75}
& \multirow{2}*{X}
& \multirow{2}*{X}
& \multirow{2}*{\textbf{6.89}}
\\
13-46-12 & & & & & \\


2014-05-14-
& \multirow{2}*{X}
& \multirow{2}*{6.30}
& \multirow{2}*{X}
& \multirow{2}*{X}
& \multirow{2}*{\textbf{5.09}}
\\
13-53-47 & & & & & \\

2014-05-14-
& \multirow{2}*{X}
& \multirow{2}*{6.15}
& \multirow{2}*{2.45}
& \multirow{2}*{X}
& \multirow{2}*{\textbf{1.83}}
\\
13-59-05 & & & & & \\

2014-06-25-
& \multirow{2}*{X}
& \multirow{2}*{3.70}
& \multirow{2}*{X}
& \multirow{2}*{6.56}
& \multirow{2}*{\textbf{3.20}}
\\
16-22-15 & & & & & \\

\hline

\end{tabular}
}
\end{center}
\label{table:robotcar_vo}
\end{table}

\begin{figure}[t!]
        \centering
        \begin{multicols}{2}
            \centering
            \includegraphics[width=0.9\columnwidth]{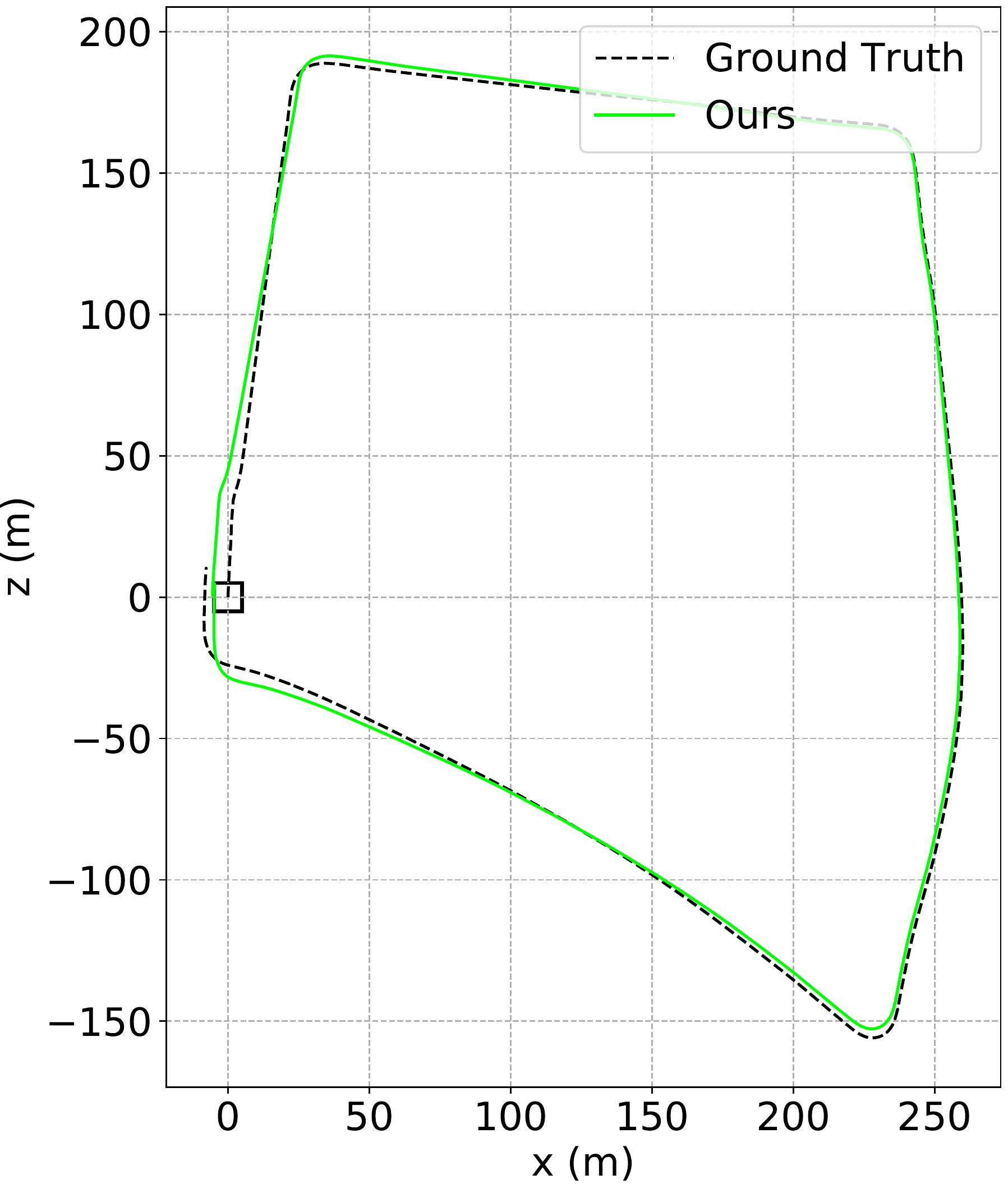}\\
            \includegraphics[width=0.9\columnwidth]{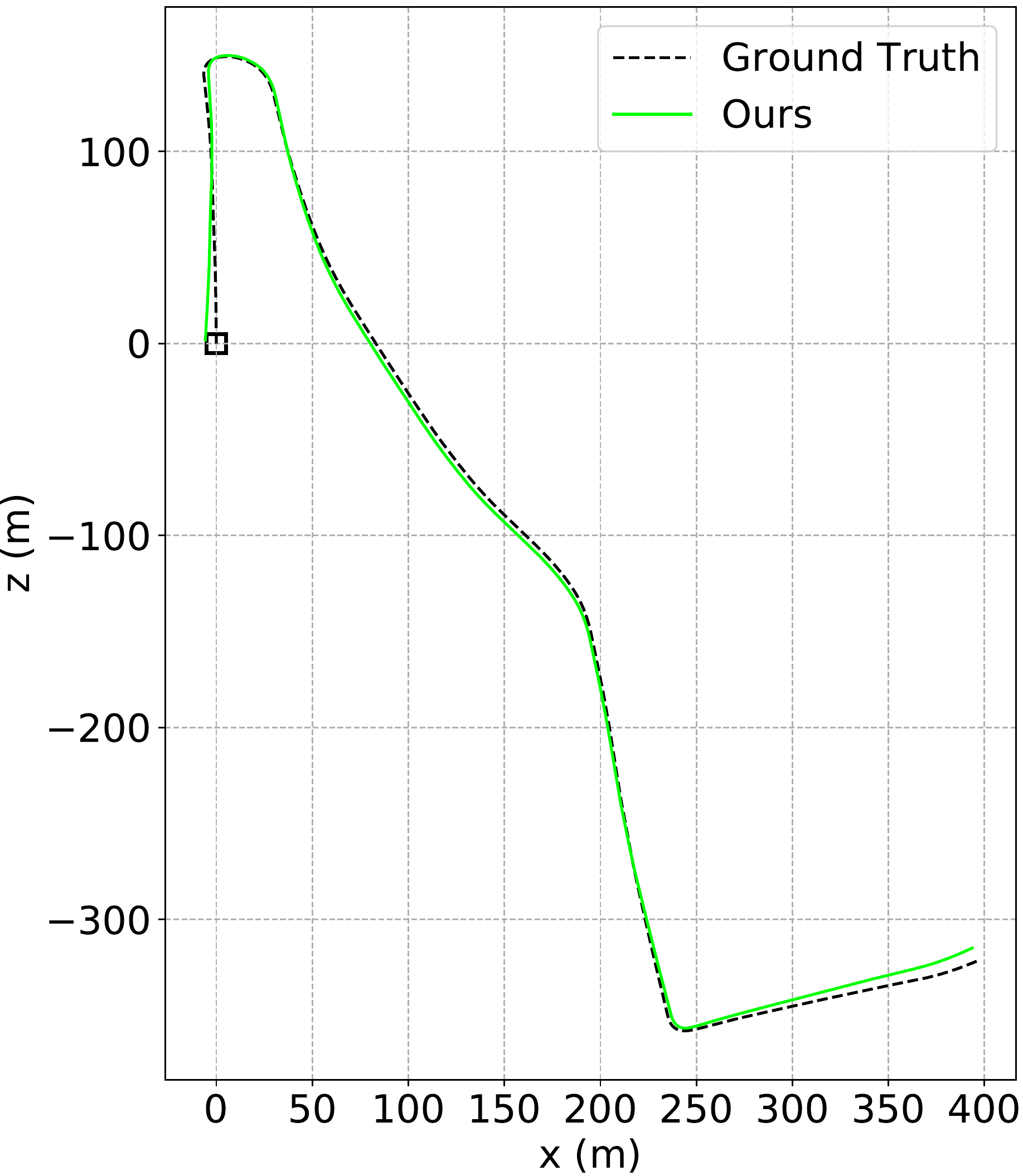}\\
        \vspace{-10pt}
        \end{multicols}
\caption{Qualitative VO results on Oxford Robotcar: (Left) 2014-05-06-12-54-54 and (Right) 2014-06-25-16-22-15. 
Note that there is in fact a loop closure in the left sequence but the "Ground truth" is not accurate enough as mentioned in the Robotcar official document.
}
\label{fig:robotcar_traj}
\end{figure}

\paragraph{Oxford Robotcar}
We also test the generalization ability of the system on Oxford Robotcar\cite{RobotCarDatasetIJRR}.
The result \footnote{The result of others are taken from \cite{loo2018cnnsvo}} is reported in \tabref{table:robotcar_vo} and illustrated in \figref{fig:robotcar_traj}.
Note that there are some overexposed frames at the middle of the sequence (e.g. \figref{fig:overexposure}), which are challenging for visual odometry/SLAM algorithms such that many algorithms listed in \tabref{table:robotcar_vo} fail to run the sequences.
However, the deep optical flow network still predicts sufficient good correspondences for pose estimation (\figref{fig:overexposure}). 
The optical network rarely fails to give sufficient good correspondences but the number of valid correspondences reflects the failure cases and constant motion model is employed in such cases.
The result shows that our system outperforms the others.
More importantly, it proves that sampling correspondence from deep optical flow is more robust than matching hand-crafted features.

\section{Ablation study} \label{sec:ablation}
\begin{table}[t!]  
\caption{
Ablation study on KITTI Odometry dataset regarding different components
}
\begin{center}
\resizebox{1\columnwidth}{!}{
\begin{tabular}{| l | l |  | c c | c c  |}
\hline
\multirow{2}*{Experiment} &  
\multirow{2}*{Variant}&
\multicolumn{2}{c|}{09} & 
\multicolumn{2}{c|}{10} \\
 {}  &
 {} &
 $t_{err}$ & $r_{err}$   &
 $t_{err}$ & $r_{err}$   \\
\hline\hline

\multicolumn{2}{|c||}{Reference Model} & 
3.45 & 0.68 & 
3.19 & 1.00  \\
\hline

Tracker &
PnP &
6.79 & 2.27 & 
6.31 & 3.75  \\
\hline

\multirow{2}*{Flow} &
Self-Flow (Offline) &
2.90 & 0.74 &
2.98 & 1.03 \\
\cline{2-6}

 &
Self-Flow (Online) &
2.07 & 0.38 &
2.54 & 0.62 \\
\hline


\multirow{2}*{Depth} &
Mono-SC &
3.45 & 0.73 &
3.63 & 1.20 \\
\cline{2-6}

{} & 
Mono. &
3.52 & 0.81 &
4.29 & 1.44 \\
\hline

\multirow{2}*{Correspondences} &
Uniform &
5.05 & 1.18 &
5.38 & 1.97 \\
\cline{2-6}

{} & 
Best-N &
4.88 & 1.06 &
4.26 & 1.83 \\
\hline

Scale &
Iterative &
3.34 & 0.63 &
3.05 & 1.07 \\
\hline

Model Sel. &
Flow &
3.71 & 0.76 &
3.57 & 1.16 \\
\hline

Img. Res. &
Full &
2.38 & 0.37 &
2.00 & 0.40 \\
\hline

\end{tabular}
}
\end{center}
\label{table:ablation_1}
\end{table}
In this section, we present an extensive ablation study (\tabref{table:ablation_1}) to understand the effect of the components proposed in this work.
We use a \textit{Reference Model} with the following settings and study the component in the following categories.
\begin{itemize}
    \item Tracker: Hybrid (E-tracker and PnP-tracker)
    \item Depth model: Trained with stereo sequences
    \item Flow model: LiteFlowNet trained from synthetic dataset
    \item Correspondence selection: Local best-K selection
    \item Scale recovery: Simple alignment
    \item Model selection: GRIC
    \item Image resolution: down-sampled size ($640\times 192$)
\end{itemize}

\subsection{Tracker}
DF-VO consists of two trackers -- E-tracker and PnP-tracker.
E-tracker is considered as the main tracker when general motion (sufficient translation) and general structure (non-planar) are assumed.
PnP-tracker is used when E-tracker fails to estimate the motion, which is introduced in \secref{sec:model_sel}.
Using E-tracker alone potentially fails when motion degeneracy or structure degeneracy happens as described in \secref{sec:epi_geo}.
Therefore, we only compare the \textit{Reference model} to the case that only PnP-tracker is used.
PnP relies on the accuracy of both depth and optical flow predictions for establishing accurate 3D-2D correspondence.
However, there is not a straightforward way to sample good depth predictions for accurate 3D-2D correspondences for 6DoF pose estimation, but the depth predictions are sufficient for 1DoF scale recovery problem in E-tracker.

\subsection{Flow model} \label{sec:ablation_flow}
\begin{table*}[t!] 
\caption{
Optical flow evaluation in KITTI 2012/2015 optical flow split. Average end-point-error (AEPE) and the percentage of pixels with error larger than 1 (Out-1) are evaluated. Non-occluded regions are evaluated. SF (Super.): supervised training on Scene Flow. KITTI (Self.): self-supervised training on KITTI. BestN: Bidirectional flow consistency thresholding is applied.
}
\begin{center}
\resizebox{0.8\textwidth}{!}{
\begin{tabular}{| l | l | c c | c c|}
\hline
\multirow{2}*{Network} &
\multirow{2}*{Dataset \& Method} &
\multicolumn{2}{c|}{KITTI 2012} & 
\multicolumn{2}{c|}{KITTI 2015} \\
 &
 &
AEPE (px) & 
Out-1 (\%) &
AEPE (px) & 
Out-1 (\%) \\
\hline\hline

LiteFlowNet &
SF (Super.) &
1.593 & 26.1\% &
4.785 & 39.6\% \\
\hline

LiteFlowNet &
SF (Super.) + KITTI (Self.) & 
1.467 & 19.7\% &
4.987 & 32.7\% \\
\hline

LiteFlowNet &
SF (Super.) + BestN &
0.478 & 7.6\% &
0.711 & 10.5\% \\
\hline

LiteFlowNet &
SF (Super.) + KITTI (Self.) + BestN & 
\textbf{0.422} & \textbf{5.7\%} &
\textbf{0.628} & \textbf{7.7\%} \\
\hline

\end{tabular}
}
\end{center}
\label{table:flow_eval}
\end{table*}
%
%

LiteFlowNet trained with synthetic data shows acceptable generalization ability from synthetic to real. 
However, there are still some regions with significantly erroneous flow predictions.
We find that with self-supervised finetuning, the model adapts better to the real world sequences and the optical flow prediction accuracy is improved (\tabref{table:flow_eval}).

\begin{figure}[t!]
            \centering
            \includegraphics[width=1\columnwidth]{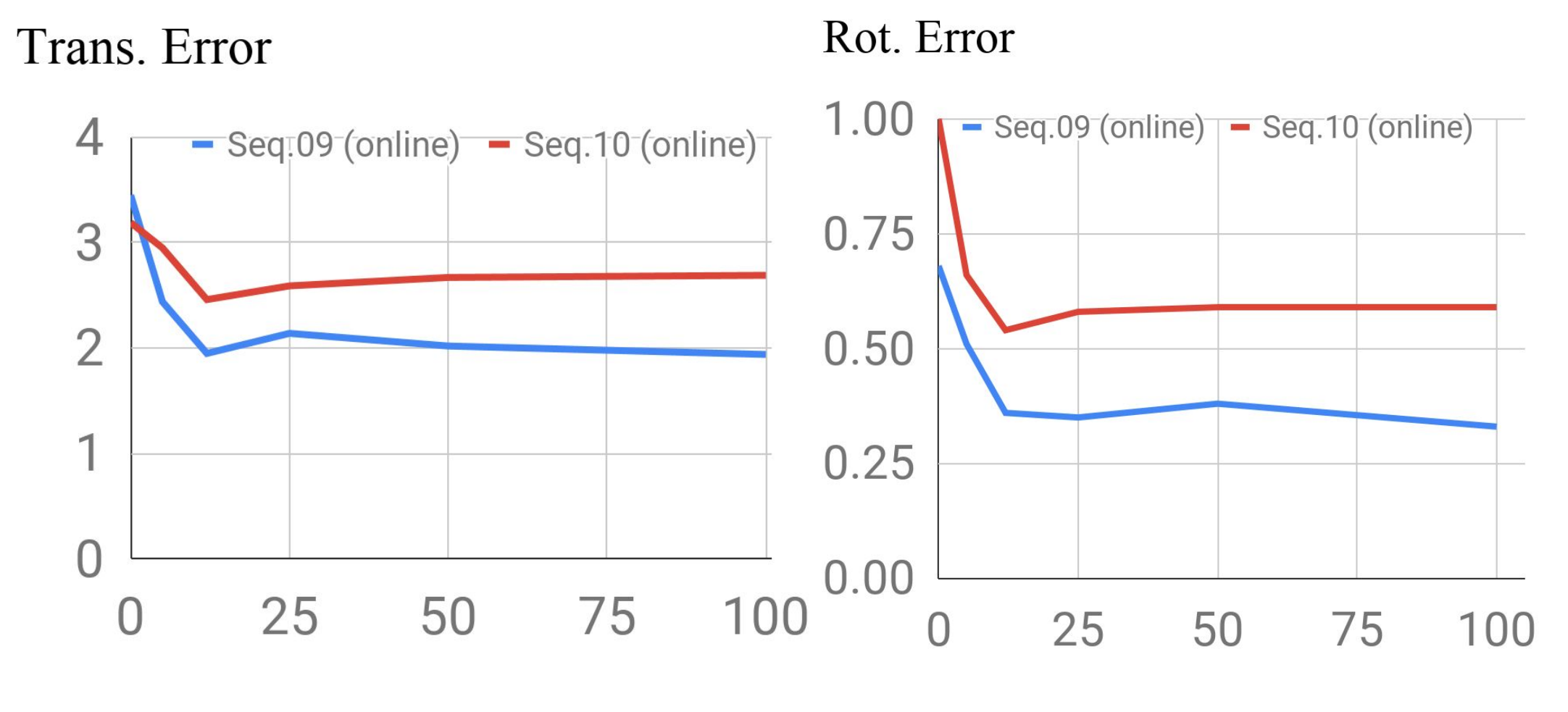}\\
\caption{Effect of self-supervised online finetuning. X-axis is the percentage of data used in the online finetuning. 
}
\label{fig:finetune_vs_vo}
\end{figure}

\paragraph{offline v.s. online}
We perform two types of self-supervised finetuning for the optical flow network.
The offline method finetunes the flow network on sequences 00-08 using monocular videos while the online method finetunes the model on-the-run for the running sequence.
We test various amounts of data for online finetuning and evaluate the corresponding odometry result. 
The relationship is shown in \figref{fig:finetune_vs_vo}. 
We can see that finetuning on a small amount of data (10\%) is sufficient for the optical flow network to adapt to unseen scenarios.

\paragraph{Flow evaluation}
We evaluate the quality of optical flows on KITTI 2012/2015, which are two benchmark dataset for optical flow evaluation.
The result is shown in \tabref{table:flow_eval}.
We can see that with self-supervised finetuning (offline), the accuracy of the flow prediction is significantly improved, especially in the percentage of outliers. 
One noticeable result is that self-supervised training increases the end-point-error in KITTI2015 from 4.785 to 4.987. 
The reason is that the self-supervised model is trained in KITTI Odometry split, which contains long driving sequences without many dynamic objects. 
However, KITTI2015 contains many dynamic objects and we observe that the error of the flow estimation on these dynamic objects are larger for the self-supervised model, which increases the average error.
On the other hand, Scene Flow model is trained in highly dynamic synthetic environments, i.e. able to estimate large flow magnitude caused by moving objects.
Moreover, the synthetic model generates artifacts in some regions when used in real-world data so there are more outliers, as shown in \tabref{table:flow_eval}.
Nevertheless, the correspondence selection module effectively removes the bad flows predicted by the self-supervised model and the overall flow accuracy is improved over the Scene Flow model.
Since better correspondences are estimated, the odometry result using Self-Flow is improved as well.

\subsection{Depth model}

Training depth models with monocular videos comes with a scale inconsistency issue \cite{bian2019depth}. 
We use an inverse depth consistency proposed in \cite{zhan2019depthnormal} to enforce the depth predictions to be consistent (\secref{sec:train_depth_consistency}).
Using a scale-consistent depth CNN for translation scale recovery helps to mitigate the scale drift issue, which usually occurs after long travelling.
Here we compare three depth models trained by different strategies.
We train two models using monocular videos. 
\textit{Mono.} model is trained without the depth consistency term while
\textit{Mono-SC} model is trained with the depth consistency term.
Models trained with monocular videos are always up-to-scale, i.e. the metric scale is unknown.
Therefore, we also train a model using stereo sequences.
Note that, the model trained with stereo sequences do not include the depth consistency term.
The predictions in stereo training are always associated with one and only one scale~i.e. real-world scale due to the constraint set by the known stereo baseline.
Therefore, no scale ambiguity/inconsistency issues exist in this training scheme.
We can see that both \textit{Reference Model (stereo)} and \textit{Mono-SC} have less $t_{err}$ and $r_{err}$ after long travelling, which is aided by the scale-consistent depth predictions.

We also explored an online adaptation scheme for the depth network. 
However, the depth network training is unstable in the online finetuning.
The scale of the depth predictions fluctuates during the training due to the scale ambiguity nature in the monocular training.

\subsection{Correspondence selection}
Since only sparse matches are required for DF-VO, a na\"ive way to extract sparse matches from dense optical flow prediction is to sample matches uniformly/randomly. 
We uniformly sampled 2000 flows to form the correspondences and it shows that the odometry result is worse than either Best-N selection or Local Best-K selection method. 
To verify the effectiveness of forward-backward flow inconsistency, which is used for correspondences selection in both Best-N selection and Local Best-K selection, we evaluate the optical flow performance with/without the selection (\tabref{table:flow_eval}).
Instead of evaluating best-N points, we alternatively set an inconsistency threshold such that only the flows with inconsistency less than $\delta_{fc}$  are evaluated.
We show that the accuracy of the selected flows is improved significantly when compared to the average result of all optical flows.

\subsection{Scale recovery}
\begin{table}[t]
\caption{
Quantitative result on KITTI tracking sequences. The RPE (m) is reported.
}
\begin{center}
\resizebox{1\columnwidth}{!}{%
\begin{tabular}{| l  | c | c | c  c | }
\hline
\multirow{2}*{Seq.} & 
{Seq. Length} & 
\multirow{2}*{ORB-SLAM2} &  
\multicolumn{2}{c|}{DF-VO (Ours)}   \\

 & 
(m) & 
 &  
Simple & 
Iterative \\

\hline\hline

2011/09/26-05 & 69.4 
& 0.053 
& 0.039
& \textbf{0.038} 
\\

2011/09/26-09 & 332.4 
& 0.061 
& 0.049
& \textbf{0.047}
\\

2011/09/26-11 & 114.0 
& 0.033 
& \textbf{0.030}
& \textbf{0.030} 
\\

2011/09/26-13 & 173.0
& 0.075 
& \textbf{0.071}
& \textbf{0.071} 
\\

2011/09/26-14 & 402.5 
& 0.101 
& \textbf{0.074}
& \textbf{0.074}
\\

2011/09/26-15 & 362.8 
& 0.087 
& 0.068 
& \textbf{0.063} 
\\

2011/09/26-18 & 51.5 
& 0.049 
& \textbf{0.014}
& 0.015 
\\

2011/09/29-04 & 254.9 
& 0.073 
& \textbf{0.040} 
& 0.044 
\\

2011/10/03-47 & 712.6 
& 0.200 
& 0.071
& \textbf{0.060} 
\\

\hline

Average & 274.8 
& 0.081 
& 0.051
& \textbf{0.049} 
\\

\hline

\end{tabular}
}
\end{center}
\label{table:kitti_tracking}
\end{table}

We propose two scale recovery methods in this work, namely simple alignment and Iterative alignment.
Simple alignment aligns the triangulated depths of the filtered optical flows and their corresponding depth predictions.
However, the filtered optical flows can fall onto dynamic object regions and the depth predictions may not be accurate.
The iterative alignment is proposed for more robust scale recovery in dynamic environments.
Only depth points and filtered optical flows that are consistent with each other are used for scale recovery.
This eliminates both bad depth predictions and optical flows of the dynamic objects.
Iterative alignment slightly improves over the simple alignment
in KITTI Odometry split, which might be because of the less dynamic scene nature of the sequences.
However, in a highly dynamic environment, like KITTI Tracking split, especially in Seq. \textit{2011/10/03-47} which is a sequence on a highway with one-third of the image occupied by moving cars, iterative scale recovery shows a better result when compared to simple alignment and works more robustly when compared to ORB-SLAM2 (\tabref{table:kitti_tracking}).

\subsection{Model selection}

Two model selection methods are proposed and tested in this work.
Flow magnitude-based method \cite{zhan2019dfvo} is straightforward but there are some potential failure cases, which is explained in \secref{sec:model_sel}.
Moreover, a flow magnitude thresholding value is required in this method, which is found empirically.
However, GRIC-based model selection is a parameter-free method, which calculates a score function for each motion model. 
It shows a more robust result when compared to the flow-based method.

\subsection{Image resolution}
Down-sampled images are used in the \textit{Reference Model} because the size is used in training deep networks.
However, simply increasing the image size to full resolution allows the optical flow network predicts more accurate correspondences thus the odometry result can be boosted easily.

\section{Conclusion} \label{sec:conclusion}
In this paper, we have presented a robust monocular VO system leveraging deep learning and geometry methods.
We explore the integration of deep predictions with classic geometry methods.
Specifically, we use optical flow and single-view depth predictions from deep networks as intermediate outputs to establish 2D-2D/3D-2D correspondences for camera pose estimation.
We show that the deep models can be trained/finetuned in a self-supervised manner and we explore the effect of various training schemes.
Depth models with consistent scale can be used for scale recovery, which mitigates the scale drift issue in most monocular VO/SLAM systems.
Instead of learning a complete VO system in an end-to-end manner, which does not perform competitively to geometry-based methods, we think integrating deep predictions with geometry gain the best from both domains.
Compared to our previous conference version \cite{zhan2019dfvo}, we robustify different components in this system and systematically evaluate the variants.
Moreover, we integrate an online adaptation scheme into the system for better adaptation ability in unseen scenarios.
A detailed ablation study is provided to verify the effectiveness of different choices in each module, including the original choices \cite{zhan2019dfvo} and the new components in this work.
With the improved system, our current version shows more robust performance, especially in highly dynamic environments.
Some prior arts \cite{yang2018dvso, tateno2017cnnslam, tang2019kp3d} show that a local optimization module is useful to further improve the VO result, which can be a future direction to improve our VO system.
Current pipeline involves a single view depth network which is less accurate than multi-view stereo (MVS) networks.
An MVS network can be considered replacing the depth network for better accuracy and possible online adaptation.

\section*{Acknowledgment}
This work was supported by the UoA Scholarship to HZ, the ARC Laureate Fellowship FL130100102 to IR and the Australian Centre of Excellence for Robotic Vision CE140100016.

{\footnotesize\bibliography{reference}}

\end{document}